%% file: main.tex
\documentclass[11pt,a4paper]{article}
\pdfoutput=1
\usepackage[hmargin=3.0cm,vmargin=3.0cm]{geometry}

\usepackage[table]{xcolor}  
\usepackage{graphicx}
\usepackage{caption,subcaption}
\usepackage{mathtools}
\usepackage{amsfonts}
\usepackage[inline]{enumitem}  
\usepackage{hyperref}
\usepackage[noabbrev]{cleveref}  
\hypersetup{
colorlinks=true,
citecolor=red
}

\usepackage[group-separator={,}]{siunitx}  
\usepackage{multicol}
\usepackage{tikz}
\usepackage{smartdiagram}
\usepackage[numbers]{natbib}  
\usepackage{authblk}            

\usepackage[space]{grffile}  

\usepackage{algorithm}
\usepackage{algpseudocode}

\newcommand\PP{\mathbb{P}}
\newcommand\RR{\mathbb{R}}
\newcommand\EE{\mathop{\mathbb{E}}}
\newcommand\zz{\mathbf{z}}
\newcommand\xx{\mathbf{x}}
\newcommand\yy{\mathbf{y}}
\newcommand\dd{\mathbf{d}}

\newcommand\manY{\mathcal{Y}}
\newcommand\manZ{\mathcal{Z}}
\newcommand\manL{\mathcal{L}}
\newcommand\tyy{\tilde{\mathbf{y}}}

\begin{document}

\title{Parametrization of stochastic inputs using generative adversarial
networks with application in geology}

\author[]{Shing Chan\footnote{Corresponding author.\\ E-mail addresses: \texttt{sc41@hw.ac.uk} (Shing Chan), \texttt{a.elsheikh@hw.ac.uk} (Ahmed H. Elsheikh).} }
\author[]{Ahmed H. Elsheikh}
\affil[]{Heriot-Watt University, United Kingdom}
\affil[]{\small School of Energy, Geoscience, Infrastructure and Society}

\maketitle

\begin{abstract}
  We investigate artificial neural networks as a parametrization tool for stochastic inputs in numerical simulations. We address parametrization from the point of view of emulating the data generating process, instead of explicitly constructing a parametric form to preserve predefined statistics of the data. This is done by training a neural network to generate samples from the data distribution using a recent deep learning technique called \emph{generative adversarial networks}. By emulating the data generating process, the relevant statistics of the data are replicated.
  The method is assessed in subsurface flow problems, where effective parametrization of underground properties such as permeability is important due to the high dimensionality and presence of high spatial correlations. We experiment with realizations of binary channelized subsurface permeability and perform uncertainty quantification and parameter estimation. Results show that the parametrization using generative adversarial networks is very effective in preserving visual realism as well as high order statistics of the flow responses, while achieving a dimensionality reduction of two orders of magnitude.
\end{abstract}

\input{intro}
\input{background}
\input{results}
\input{conclusion}

\bibliographystyle{plainnat}
\bibliography{biblio}

\pagebreak

\end{document}

%% file: intro.tex
\section{Introduction}

Many problem scenarios such as uncertainty quantification and parameter estimation
involve the solution of partial differential equations with a
stochastic input. Input in this context is understood as any parameter
that affects the system response, e.g. the conductivity parameter in the heat
equation.
This is because in many real applications, some properties of the system are
uncertain or simply unknown. The general approach is to use a probabilistic
framework where we represent such uncertainties as random variables with a
predefined distribution. In some cases where both the
distribution and the forward map are trivial, a closed-form solution is
possible; however this is very rarely the case. Often in practice, we can
only resort to a brute-force approach where we draw several realizations of
the random variables and fully solve the partial differential equations for
each realization in an effort to estimate distributions or bounds of the
system's response. 
This approach suffers from slow convergence and the need to perform
a large number of forward simulations, which led to the development of several
methods to reduce the computational burden of this task.

A straightforward solution is to reduce the computational cost of the forward map itself -- numerous methods exist in this direction.
Another different direction consists of reducing or refining the search
space or distribution of the random variables, for example by regularization
or parametrization, thus reducing the number of simulations required.
Parametrization is specially useful in problems where the number of random variables is huge but the variables are highly redundant and correlated. This is generally the case in subsurface flow problems: 
Complete prior knowledge of subsurface properties (e.g. porosity or permeability) is impossible, yet is very influential in the flow responses. 
At the same time, accurate flow modeling often requires the use of extremely large simulation grids. When the subsurface property is discretized, the number of free variables is naively associated with the number of grid cells. The random variables thus obtained are hardly independent, whose assumption during the modeling leads to unnecessary computations. The goal of parametrization is to discover statistical relationships between the random variables in order to obtain a reduced and more effective representation.

The importance of parametrization in subsurface simulations resulted in a variety of methods in the literature including 
zonation \citep{jacquard1965,jahns1966} and zonation-based methods
\citep{bissell1994calculating,chavent1998indicator,grimstad2001scale,grimstad2002identification,grimstad2003adaptive,aanonsen2005efficient},
PCA-based methods~\citep{gavalas1976reservoir,oliver1996multiple,reynolds1996reparameterization,sarma2008kernel,ma2011kernel,vo2016regularized},
SVD-based methods~\citep{shirangi2014history,shirangi2016improved,tavakoli2009history,tavakoli2011monte},
discrete wavelet transform~\citep{mallat1989multiresolution,lu2000multiresolution,sahni2005multiresolution},
discrete cosine transform~\citep{jafarpour2007,jafarpour2009,jafarpour2010},
level set methods~\citep{moreno2007stochastic,dorn2008,chang20108011}, and
dictionary learning~\citep{khaninezhad2012sparse1,khaninezhad2012sparse2}.
Many methods begin by proposing parametric forms for the
random vector to be modeled which are then explicitly fitted to preserve
certain chosen statistics.
In the process, many methods inevitably adopt some simplifying assumptions, either on the parametric form to be employed or the statistics to be reproduced, which are necessary for the method to be actually feasible.
In this work, we consider the use of neural networks for \emph{both} parametrization of the random vector \emph{and} definition of its relevant statistics.
This is motivated by recent advances in the field of machine learning and the high expressive power of neural networks that makes them one of the most flexible forms of parametrization.

The idea is to obtain a parametrization by emulating the data generating process.
We seek to construct a function called the generator -- in this case, a neural network -- that takes a low-dimensional vector as input (the reduced representation), and aims to output a realization of the target random vector. The low-dimensional vector is assumed to come from an easy-to-sample distribution, e.g. a multivariate normal or an uniform distribution, and is what provides the element of stochasticity. Generating a new realization then only requires sampling the low-dimensional vector and a forward pass of the generator network.
The neural network is trained using a dataset of prior realizations that inform the patterns and variability of the random vector
(e.g. geological realizations from a database or from multipoint geostatistical simulations~\citep{strebelle2001reservoir,mariethoz2014multiple}).

The missing component in the description above is the definition of an objective function to actually train such generator; in particular, how do we quantify the discrepancy between generated samples and actual samples? This is resolved using a recent technique in machine learning called \emph{generative adversarial networks}~\citep{goodfellow2014generative} (GAN).
The idea in GAN is to let a second classifier neural network, called the discriminator, define the objective function. The discriminator takes the role of an adversary against the generator where both are trained alternately in a minmax game: the discriminator is trained to maximize a classification performance where it needs to distinguish between ``fake'' (from the generator) and ``real'' (from the dataset) samples, while the generator is trained to minimize it. Hence, the generator is iteratively trained to generate good realizations in order to fool the discriminator, while the discriminator is in turn iteratively trained to improve its ability to classify correctly.
In the equilibrium of the game, the generator effectively learns to generate plausible samples, and the discriminator regards all samples as equally plausible (coin toss scenario). 

The benefit of this approach is that we do not need to manually specify which statistics of the data need to be preserved, instead we let the discriminator network implicitly learn the relevant statistics.
We can see that the high expressive power of neural networks is leveraged in two aspects: on one hand, the expressive power of neural networks allows the effective parametrization (generator) of complex realizations; on the other hand, it allows the discriminator to learn the complex statistics of the data.

In this work, we parametrize binary channelized permeability models based on
the classical Strebelle training image~\citep{strebelle2001reservoir}, a
benchmark problem that is often employed for assessing parametrization methods
due to the difficulties in crisply preserving the channels.
To assess the method, we consider uncertainty propagation in subsurface flow problems for a large number of realizations of the permeability and compare the statistics in their flow responses. We also perform parameter estimation using natural evolution strategies~\citep{wierstra2008natural,wierstra2014natural}, a general black-box optimization method that is suitable for the obtained reduced representation.
Finally, we discuss training difficulties of GAN encountered during our implementation -- in particular, limitations in small datasets, and inherent issues of the standard formulation of GAN~\citep{goodfellow2014generative}.
This work is an extension of our preliminary work
in~\citep{chan2017parametrization}. There is a growing number of recent works in
geology-related fields where GAN methods have been studied.
In~\citep{mosser2017reconstruction,mosser2017stochastic},
GAN is used to generate images of porous media for image reconstruction.
In~\citep{mosser2018stochastic}, GAN is used in seismic inversion.
In~\citep{laloy2018training,zhang2019generating}, GAN is used to generate geological
models for parameter estimation.
In~\citep{dupont2018generating,mosser2018conditioning}, GAN is used
to generate conditional realizations. In this work, we provide further
assessments and focus mainly on the capabilities of GAN as a parametrization
tool to preserve high order flow statistics and visual realism.

The rest of this chapter is organized as follows: In~\Cref{sec:ch3background}, we briefly
describe convolutional neural networks -- a widely used architecture
in modern neural networks -- and the method of generative
adversarial networks. \Cref{sec:ch3results} contains our numerical results for
uncertainty quantification and parameter estimation experiments. In~\Cref{sec:practical},
we provide a discussion for practical implementation. We draw our conclusions in~\Cref{sec:ch3conclusion}.

%% file: background.tex
\section{Background}\label{sec:ch3background}

\subsection{Convolutional neural networks} \label{sec:cnn}
\def\layersep{2.5cm}

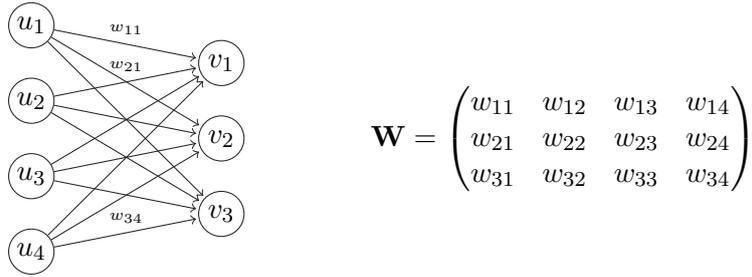
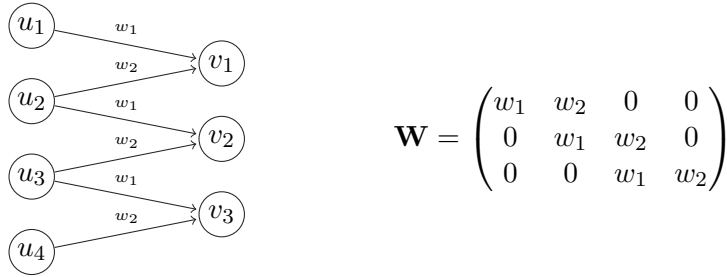
\begin{figure}
  \begin{subfigure}{\textwidth}
  \centering
  \begin{tikzpicture}[shorten >=1pt,->,draw=black!85, node distance=\layersep]
    \tikzstyle{every pin edge}=[<-,shorten <=1pt]
    \tikzstyle{neuron}=[circle,draw,minimum size=17pt,inner sep=0pt]
    \tikzstyle{annot} = [text width=4em, text centered]

    \foreach \name / \y in {1,...,4}
    \node[neuron] (I-\name) at (0,-\y) {$u_{\y}$};

    \foreach \name / \y in {1,...,3}
    \path[yshift=-0.5cm]
    node[neuron] (H-\name) at (\layersep,-\y cm) {$v_{\y}$};

    \foreach \source in {1,...,4}
    \foreach \dest in {1,...,3}
    \path (I-\source) edge (H-\dest);

    \path (I-1) -- (H-1) node[midway,above]{\tiny $w_{11}$} ;
    \path (I-1) -- (H-2) node[midway,above]{\tiny $w_{21}$} ;
    \path (I-4) -- (H-3) node[midway,above]{\tiny $w_{34}$} ;

    \node (matrix) at (7,-2.5) {
      $\mathbf W = 
      \begin{pmatrix}
        w_{11} & w_{12} & w_{13} & w_{14} \\
        w_{21} & w_{22} & w_{23} & w_{24} \\
        w_{31} & w_{32} & w_{33} & w_{34}
      \end{pmatrix}$

    };
  \end{tikzpicture}
  \caption{A fully connected layer.}
  \label{fig:fc}
  \end{subfigure}
  \vspace{1em}

  \begin{subfigure}{\textwidth}
  \centering
  \begin{tikzpicture}[shorten >=1pt,->,draw=black!85, node distance=\layersep]
    \tikzstyle{every pin edge}=[<-,shorten <=1pt]
    \tikzstyle{neuron}=[circle,draw,minimum size=17pt,inner sep=0pt]
    \tikzstyle{annot} = [text width=4em, text centered]

    \foreach \name / \y in {1,...,4}
    \node[neuron] (I-\name) at (0,-\y) {$u_{\y}$};

    \foreach \name / \y in {1,...,3}
    \path[yshift=-0.5cm]
    node[neuron] (H-\name) at (\layersep,-\y cm) {$v_{\y}$};

    \foreach \dest in {1,...,3}{
    \pgfmathtruncatemacro{\destplusone}{\dest + 1};
    \path (I-\dest) edge node[midway,above]{\tiny $w_1$} (H-\dest);
    \path (I-\destplusone) edge node[midway,above]{\tiny $w_2$} (H-\dest);
    }

    \node (matrix) at (7,-2.5) {
      $\mathbf W = 
      \begin{pmatrix}
        w_{1} & w_{2} & 0 & 0 \\
        0 & w_{1} & w_{2} & 0 \\
        0 & 0 & w_{1} & w_{2}
      \end{pmatrix}$

    };
  \end{tikzpicture}
  \caption{A convolutional layer.}
  \label{fig:conv}
  \end{subfigure}
  \caption{Transformation matrix of a fully connected layer (a), and of a
    convolutional layer (b). In this example, the convolutional layer has only
    $2$ free weights, whereas the fully connected layer has $12$ free weights.}
  \label{fig:ch3nn}
\end{figure}
A (feedforward) neural network is a composition of functions
$f(\xx)=f^L(f^{L-1}(\cdots(f^1(\xx))))$ where each function $f^l(\mathbf x)$,
called a layer, is of the form $\sigma_l(\mathbf W_l\mathbf x + \mathbf b_l)$, i.e. an affine
transform followed by a component-wise non-linearity.
The choice of the number of layers $L$, the non-linear functions $\sigma_l$,
and the sizes of $\mathbf W_l, \mathbf b_l$ are part of the \emph{architecture design} process,
which is largely problem-dependent and based on heuristics and domain knowledge.
Modern architectures use non-linearities such as rectifier linear units (ReLU, $\sigma(x) = x^{+} = \max(0,x)$),
leaky rectifier linear units (leaky ReLU, $\sigma(x) = x^{+} + 0.01x^{-}$),
$\tanh$,  sigmoid, and others;
and can have as much as $100$ layers~\citep{he2016deep,simonyan2014very}.
After an architecture is assumed, the weights of $\mathbf W_l, \mathbf b_l$ are obtained
by optimization of an objective.

A major architectural choice that led to huge advances in computer vision is the
use of \emph{convolutional layers}~\citep{fukushima1982neocognitron}. An example
of a convolution is the following: Let $\mathbf u = (u_1,\cdots,u_m)$ be an
input vector and $\mathbf k = (w_1, w_2)$ be a \emph{filter}. The output of
convolving the filter $\mathbf k$ on $\mathbf u$ is
$\mathbf v = (v_1,\cdots,v_p)$ where $v_i = w_1u_i + w_2u_{i+1}$
(using stride $1$). 
The operation is illustrated in~\Cref{fig:ch3nn}: In a traditional fully
connected layer, the associated matrix is dense and all its weights need to be
determined in optimization. In a convolutional layer, the connections are
constrained in such a way that each output component only depends on a
neighborhood of the input, and as a result the
associated matrix is a sparse diagonal-constant matrix, thus obtaining a huge reduction in the number of free weights.
Note that in practice several stacks of convolution layers are used, so
the full connectivity can be recovered if necessary.
On top of the reduction, another benefit comes from the inherent
regularization that this operator imposes that is often useful in
applications where there is a spatial or temporal extent and the assumption
of data locality is valid -- informally, closer events in the
spatial/temporal extent tend to be correlated (e.g. in natural images and
speech).

The convolution as described above has a downscalng effect, i.e. the output
size is always smaller or equal to the input size, which can be controlled by
the filter stride. A classifier neural network typically consists of a series of
convolutional layers that successively downscale an image to a single number
(binary classification) or a vector of numbers (multiclass classification).
In the case of decoders and generative networks, we wish to achieve the
opposite effect to get an output that is larger than the input (e.g. to
reconstruct an image from a compressed code).
This can be achieved by simply \emph{transposing} the convolutions: Using
the example in~\Cref{fig:conv}, to convert from $\mathbf v$ to $\mathbf u$,
we can consider weight matrices of the form $\mathbf W^\top$, i.e. the
transpose of the convolution matrix from $\mathbf u$ to $\mathbf v$. Modern
decoders and generators consist of a series of transposed
convolutions that successively upscale a small vector to a large output
array, e.g. corresponding to an image or audio.

The operations and properties discussed so far extend naturally to 2D and 3D
arrays. For a 2D or 3D input array, the filter is also a 2D or 3D array,
respectively. Note however that in the 3D case, the filter array is such that
the depth (perpendicular to the spatial extent) is always equal to the depth of the
3D input array, therefore the output is always a 2D array, and the striding is
done in the spatial extent (width and height). On the other hand, we allow the
application of multiple filters to the same input, thereby producing a 3D
output array if required, consisting of the stack of multiple convolution outputs. This way of
operating with convolutions is inherited from image processing: Color images are
3D arrays consisting of three 2D arrays indicating the red, green, and
blue intensities (in RGB format). Image filtering normally operates on
all three values, e.g. the greyscale filter is
$v_{ij} = 0.299\cdot u_{ij,\mathrm{red}} + 0.587\cdot u_{ij,\mathrm{green}} + 0.114\cdot u_{ij,\mathrm{blue}}$.
The output of a convolution filter is also called a feature map.

\begin{figure}
  \centering
  \includegraphics[width=.5\textwidth]{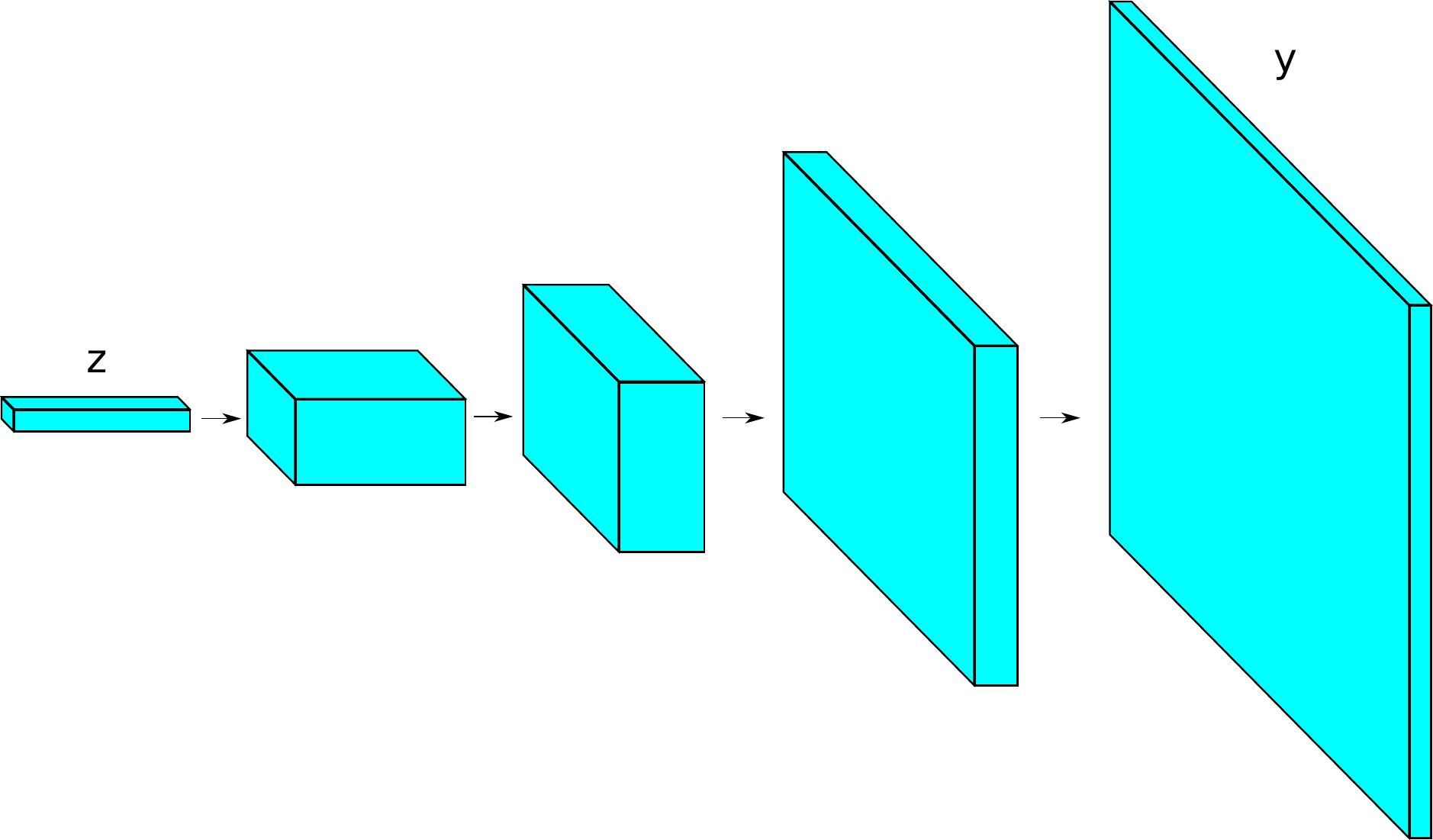}
  \caption{Illustration of a typical pyramid architecture used in generator networks.}
  \label{fig:cnn}
\end{figure}

Finally, we show in~\Cref{fig:cnn} a popular pyramid architecture used in
generator networks~\citep{radford2015unsupervised} for image synthesis. The blocks shown
represent the state shapes (stack of feature maps) as the
input vector is passed through the network. The input vector $\zz$ is first
treated as a ``1-pixel image'' (with many ``feature maps''). 
The blocks are subsequently expanded in the spatial extent (width and height)
while thinned in depth:
A series of transposed convolutions is used to upsample the spatial extent
until reaching the desired size; at the same time, the number of
convolution filters is initially large, but it is 
subsequently reduced in the following layers. For classifier networks, usually the inverted architecture
is used where the transposed convolutions are replaced with normal
convolutions.
Further notes on modern convolutional neural networks can be found
in~\citep{cs231n}.

\subsection{Generative adversarial networks}
Let $\zz \sim p_z$, $\yy \sim \PP_y$, where $p_z$ is a known, easy-to-sample
distribution (e.g. multivariate normal or uniform distribution), and $\PP_y$ is the unknown
distribution of interest (e.g. the distribution of all possible geomodels in a
particular zone). The distribution $\PP_y$ is only known through realizations
$\{\yy_1, \yy_2, \cdots, \yy_n\}$ (e.g. realizations provided by multipoint
geostatistical simulations).
Let $G_{\theta}\colon\manZ\to\manY$ be a neural network -- called the
generator -- parametrized by weights $\theta$ to be determined. Given $p_z$ fixed, this neural
network induces a distribution $G_\theta(\zz)\sim\PP_{\theta}$ that depends on $\theta$, and whose
explicit form is complicated or intractable (since neural networks
contain multiple non-linearities). On the other hand, sampling from this
distribution is easy since it only involves sampling $\zz$ and a forward evaluation
of $G_\theta$.
The goal is to find $\theta$ such that $\PP_{\theta} = \PP_y$.

Generative adversarial networks (GAN)~\citep{goodfellow2014generative} approach
this problem by considering a second classifier neural network -- called the
discriminator -- to classify between ``fake'' samples (generated by the
generator) and ``real'' samples (coming from the dataset of realizations). Let
$D_\psi\colon\manY\to [0,1]$ be the discriminator network parametrized by weights $\psi$ to be
determined. The training of the generator and discriminator uses the following loss function:

\begin{equation}
  \label{eq:ch3ganloss}
\manL(\psi,\theta)  \coloneqq \EE_{\yy\sim\PP_y}{\log D_{\psi}(\yy)} + \EE_{\tilde{\yy}\sim\PP_\theta}{\log(1-D_{\psi}(\tilde{\yy}))} 
\end{equation}
where $\tilde{\yy} = G_\theta(\zz)\sim\PP_\theta$. In effect, this loss
is the classification score of the discriminator, therefore we train $D_\psi$
to maximize this function, and $G_\theta$ to minimize it:
\begin{equation}
  \label{eq:ch3minmax}
  \min_\theta \max_\psi {\manL(\psi,\theta)}
\end{equation}

In practice, optimization of this minmax game is done alternately using some
variant of stochastic gradient descent, where the gradient can be obtained using
automatic differentiation algorithms. 
In the infinite capacity setting, this optimization amounts to minimizing the Jensen-Shannon
divergence between $\PP_y$ and $\PP_\theta$~\citep{goodfellow2014generative}. 
Equilibrium of the game occurs when $\PP_y=\PP_{\theta}$ and
$D_{\psi}=\frac{1}{2}$ in the support of $\PP_y$ (coin toss scenario).

\subsubsection{Wasserstein GAN}
In practice, optimization of the minmax game~\eqref{eq:ch3minmax} is known to be
very unstable, prompting numerous works to understand and address this
issue~\citep{radford2015unsupervised,salimans2016improved,berthelot2017began,qi2017loss,kodali2017train,arjovsky2017towards,arjovsky2017wasserstein,gulrajani2017improved}.
One recent advance is to use the Wasserstein formulation of
GAN (WGAN)~\citep{arjovsky2017wasserstein,gulrajani2017improved}.
This formulation proposes the objective function

\begin{equation}
  \label{eq:wganloss}
\manL(\psi,\theta)  \coloneqq \EE_{\yy\sim\PP_y}{D_{\psi}(\yy)} - \EE_{\tilde{\yy}\sim\PP_\theta}{D_{\psi}(\tilde{\yy})} 
\end{equation}
and a constraint in the search space of $D_\psi$,
\begin{equation}
  \label{eq:ch3minmax2}
  \min_\theta \max_{\psi:{D_\psi\in\mathcal{D}}} {\manL(\psi,\theta)}
\end{equation}
where now $D_\psi\colon\manY\to\RR$ and $\mathcal D$ is the set of
$1$-Lipschitz functions. This constraint can be loosely enforced by constraining
the weights $\psi$ to a compact space, e.g. by clipping the values of the
weights in an interval $[-c,c]$. In practice, $\mathcal D$ is a set of
$k$-Lipschitz functions for a constant $k$ that is irrelevant for optimization.
Although the modifications in~\Cref{eq:wganloss,eq:ch3minmax2}
over~\Cref{eq:ch3ganloss,eq:ch3minmax} seem trivial, the derivation of this
formulation is rather involved and can be found in~\citep{arjovsky2017wasserstein}. In essence,
this formulation aims to minimize the \emph{Wasserstein distance} between two
distributions, instead of the Jensen-Shannon divergence. Here we only
highlight important consequences of this formulation:
\begin{itemize}
\item Available meaningful loss metric. This is because
  \begin{equation}\label{eq:wasserstein}
  W(\PP_y,\PP_\theta) \approx \max_{\psi:{D_\psi\in\mathcal{D}}} {\manL(\psi,\theta)}
  \end{equation}
  where $W$ denotes the Wasserstein distance.
  \item Better stability. In particular, mode collapse is drastically reduced (see~\Cref{sec:importance}).
  \item Robustness to architectural choices and optimization parameters.
\end{itemize}
We experimentally verify these points in~\Cref{sec:importance} and discuss
their implications for our current application.

A pseudo-code of the training process is shown in~\Cref{algo:wgan}.
Note that $D$ is trained multiple times ($n_D$) per each iteration of $G$. This
is to keep $D$ near optimality so that the Wasserstein estimate
in~\Cref{eq:wasserstein} is accurate before every update of $G$. We also note that
even though we show a simple gradient ascent/descent in the update steps (lines
6 and 11), it is more common to use update schemes such as
RMSProp~\citep{tieleman2012lecture} and Adam~\citep{kingma2014adam} that are
better suited for neural network optimization.

\begin{algorithm}
  \caption{The WGAN algorithm}\label{algo:wgan}
  \begin{algorithmic}[1]
  \Require{$n_D$ iterations of $D$ per iteration of $G$, initial guesses $\theta_{\mathrm{init}},\psi_{\mathrm{init}}$, step size $\eta$, batch size $m$, clipping interval $c$.}
    \While{$\theta$ has not converged}

    \Comment{Train $D$}
      \For{$t = 1, ..., n_{D}$}
        \State Sample $\{\zz_1,\cdots,\zz_m\} \sim \PP_z$ to get $\{\tilde{\yy}_1,\cdots,\tilde{\yy}_m\},\; \tilde{\yy}_i = G_\theta(\zz_i)$
        \State Sample $\{\yy_1,\cdots,\yy_m\} \sim \PP_y$ (draw a subset of the dataset)
        \State $\nabla_\psi\manL(\psi,\theta) \gets \nabla_\psi \left[\frac{1}{m}\sum_{i=1}^m D_\psi(\yy_i) - \frac{1}{m} \sum_{i=1}^m D_\psi(\tyy_i) \right]$
        \State $\psi \gets \psi + \eta \nabla_\psi\manL(\psi,\theta)$
        \State $\psi \gets \text{clip}(\psi, -c, c) $
      \EndFor

      \Comment{Train $G$}
      \State Sample $\{\zz_1,\cdots,\zz_m\} \sim \PP_z$ to get $\{\tilde{\yy}_1,\cdots,\tilde{\yy}_m\},\; \tilde{\yy}_i = G_\theta(\zz_i)$
      \State $\nabla_\theta\manL(\psi,\theta) \gets -\nabla_\theta \frac{1}{m} \sum_{i=1}^m D_\psi(\tyy_i)$ 
      \State $\theta \gets \theta - \eta \nabla_\theta\manL(\psi,\theta)$
    \EndWhile
\end{algorithmic}
\end{algorithm}  

%% file: results.tex
\section{Numerical experiments} \label{sec:ch3results}
We perform parametrization of unconditional and conditional realizations
(point conditioning) of size $64\times 64$ of a binary channelized
permeability by training GAN using 1000 prior realizations. The training
image is the benchmark image by Strebelle~\citep{strebelle2001reservoir}
containing meandering left-to-right channels.
The conditioning is done at 16 locations, summarized in
Table~\ref{table:hdd}, containing 13 locations of high permeability (channel
material) and 3 locations of low permeability (background material). The
channel has a log-permeability of $1$ and the background has a
log-permeability of $0$, however for the purpose of training GAN we shall
use $-1$ to denote the background, and restore the value to $0$ for the
flow simulations.
The prior realizations are generated using the \emph{snesim} algorithm~\citep{strebelle2001reservoir}.
Examples of such realizations are shown in the top rows of~\Cref{fig:uncond_samples,fig:cond_samples}.

\begin{table}
  \centering
  \begin{tabular}{r| c c c c}
           & $j=12$ & $j=25$ & $j=38$ & $j=51$ \\
    \hline
    $i=12$ & $1$ & $1$ & $1$ & $1$ \\
    $i=25$ & $1$ & $1$ & $0$ & $0$ \\
    $i=38$ & $1$ & $0$ & $1$ & $1$ \\
    $i=51$ & $1$ & $1$ & $1$ & $1$ \\
  \end{tabular}
  \caption{Point conditioning at $16$ locations, indicated by cell indices
    $(i,j)$, regularly distributed across the domain.}
  \label{table:hdd}
\end{table}

\subsection{Implementation}

\begin{figure}\centering
  \begin{minipage}[b]{.45\textwidth}
    \includegraphics[width=\textwidth]{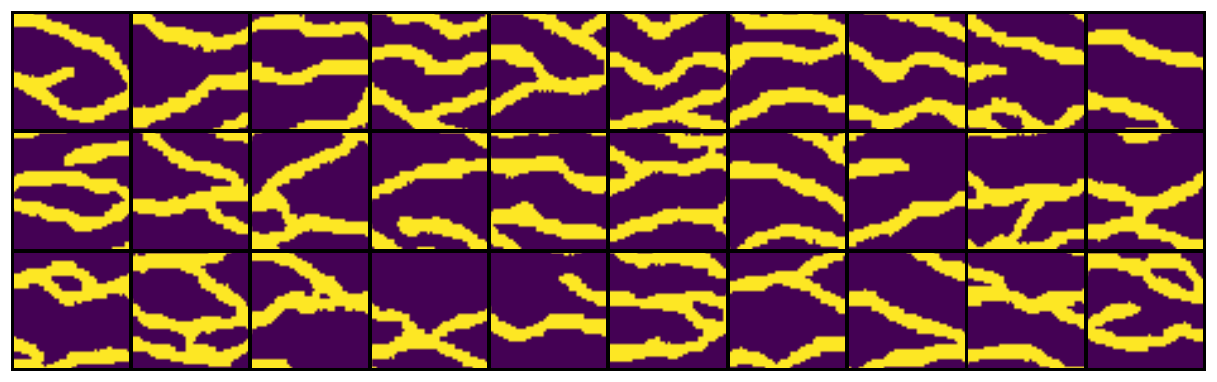}
    \vspace{-20pt} \captionof*{subfigure}{Reference} \vspace{4pt}
    \includegraphics[width=\textwidth]{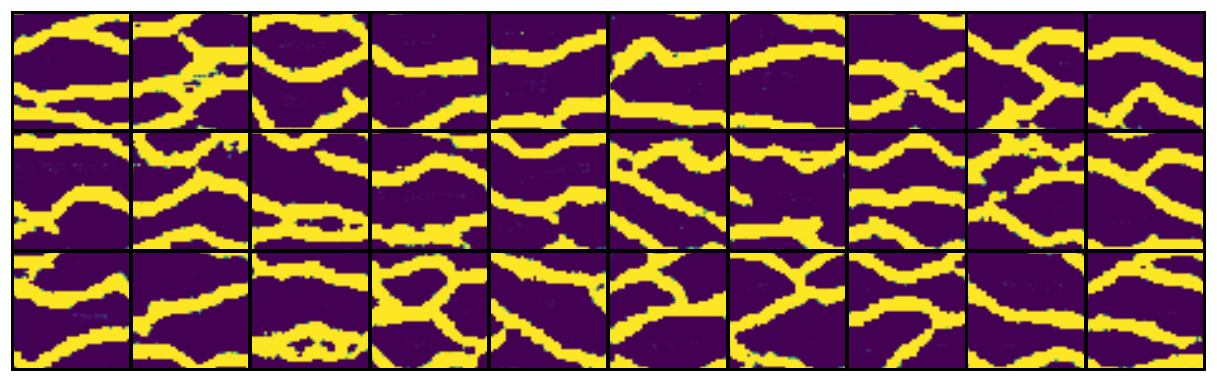}
    \vspace{-20pt} \captionof*{subfigure}{WGAN} \vspace{4pt}
    \includegraphics[width=\textwidth]{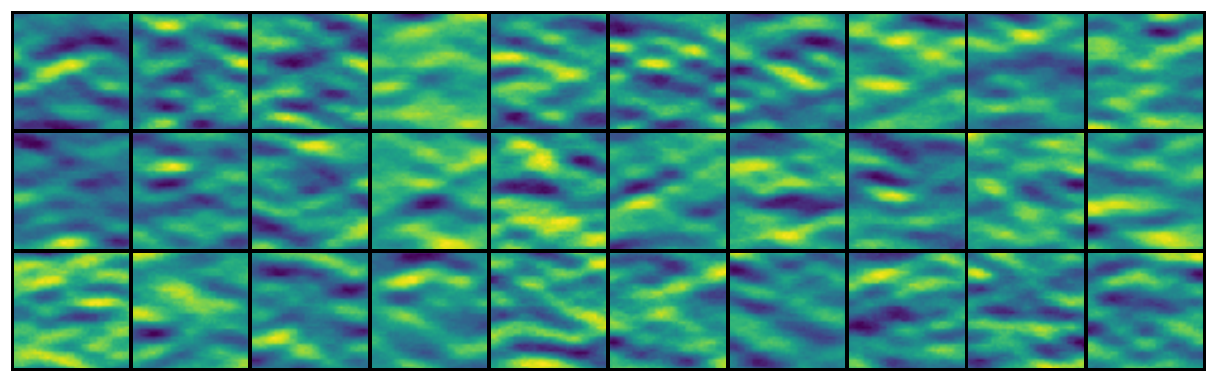}
    \vspace{-20pt} \captionof*{subfigure}{PCA} 
    \captionof{figure}{Unconditional realizations}
    \label{fig:uncond_samples}
  \end{minipage}
  \begin{minipage}[b]{.45\textwidth}
    \includegraphics[width=\textwidth]{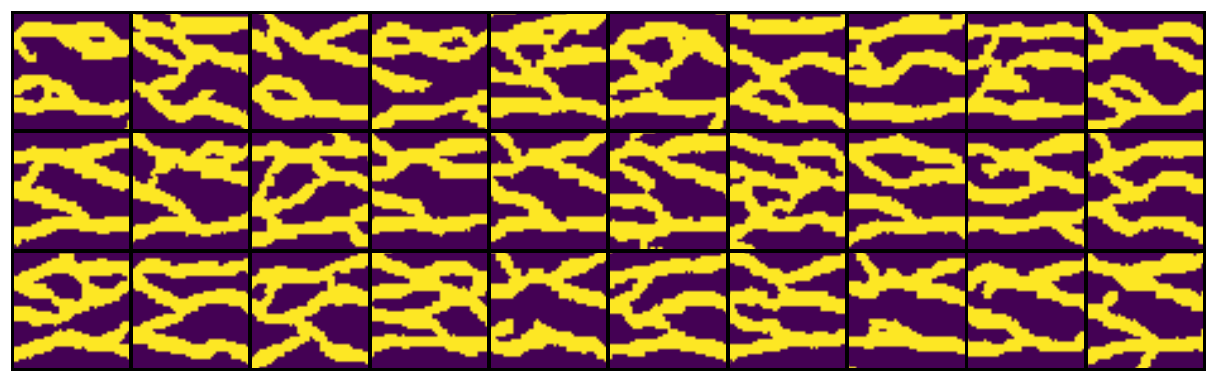}
    \vspace{-20pt} \captionof*{subfigure}{Reference} \vspace{4pt}
    \includegraphics[width=\textwidth]{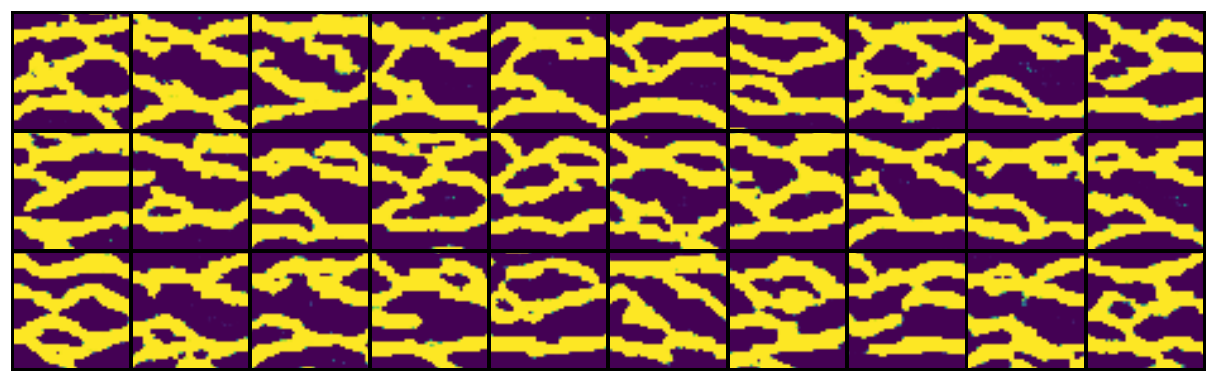}
    \vspace{-20pt} \captionof*{subfigure}{WGAN} \vspace{4pt}
    \includegraphics[width=\textwidth]{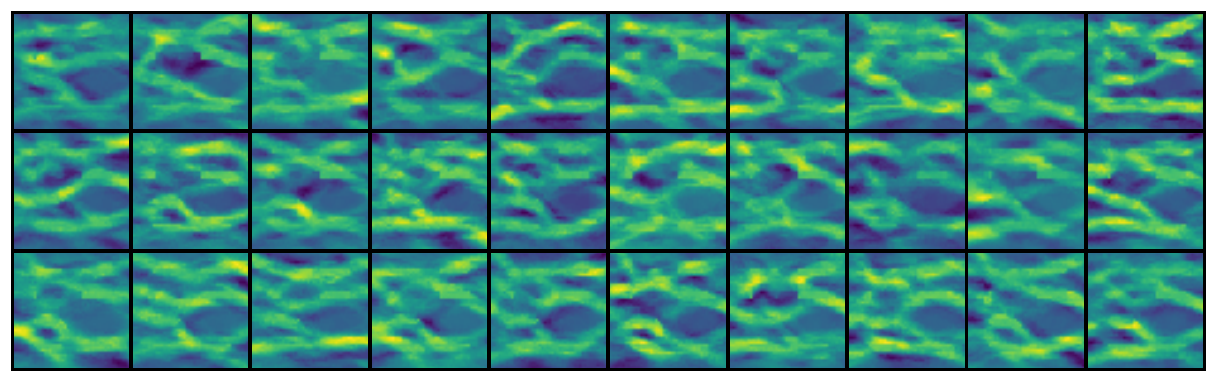}
    \vspace{-20pt} \captionof*{subfigure}{PCA} 
    \captionof{figure}{Conditional realizations}
    \label{fig:cond_samples}
  \end{minipage}
\end{figure}

\begin{figure} \centering
  \begin{subfigure}{\textwidth}
    \includegraphics[width=\textwidth]{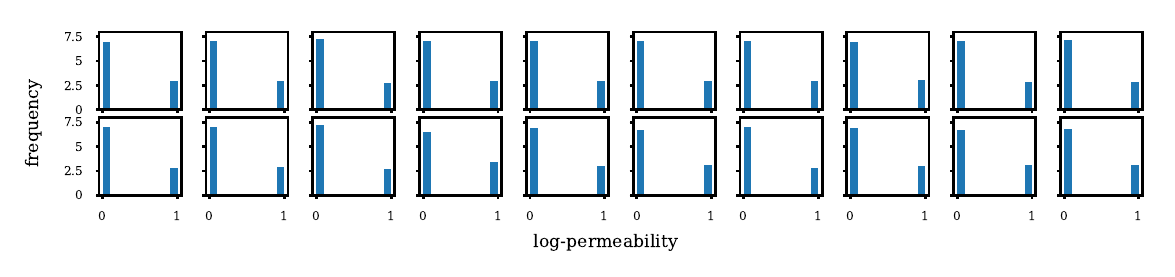}
    \vspace{-2em}\caption{Unconditional case}\vspace{0.5em}
  \end{subfigure}
  \begin{subfigure}{\textwidth}
    \includegraphics[width=\textwidth]{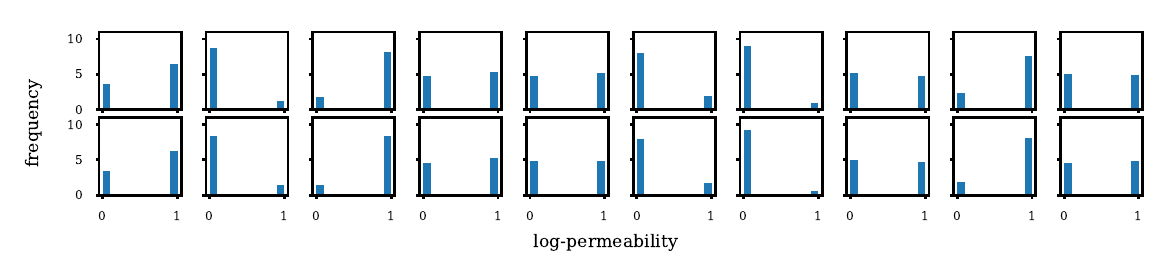}
    \vspace{-2em}\caption{Conditional case}
  \end{subfigure}
  \caption{Histogram of permeability at $10$ random locations based on
    snesim (first row) and WGAN (second row) realizations.}
  \label{fig:permhist}
\end{figure}

We train separate WGAN models for unconditional and conditional realizations.
The architecture is the pyramid structure as described in ~\Cref{sec:cnn} and shown in~\Cref{fig:cnn}:
For the generator, the input array is initially upscaled to size $4\times4$ in
the spatial extent, and the initial number of feature maps is $512$. This ``block'' is
successively upscaled in the spatial extent and reduced in the number of feature
maps both by a factor of 2, until the spatial extent reaches size $32\times32$. A final
transposed convolution upscales the block to the output size of $64\times64$. The non-linearities
are ReLUs for all layers except the last layer where we use $\tanh(\cdot)$
(so that the output is bounded in $[-1,1]$ -- as noted above, we denote the background with $-1$ and the channels with $1$).
The discriminator architecture is a mirror version of the generator
architecture, where the initial number of feature maps is $8$. The block is
successively downscaled in the spatial extent and increased in the number of
feature maps by a factor of 2, until the spatial extent reaches size $4\times4$. A
final convolutional filter reduces this block to a single real value. Note
that the size of the discriminator (in terms of total number of weights) is
$1/8$ times smaller than the generator, which is justified below
in~\Cref{sec:netsize}. All layers except the last use leaky ReLUs. The last
layer does not use a non-linearity.
We use $\mathbf z \sim \mathcal N(\mathbf 0,\mathbf I)$ of dimension $30$. This
was chosen using principal component analysis as a rule of thumb: to retain
$75\%$ of the energy, $54$ and $94$ eigencomponents are required in the
unconditional and conditional cases, respectively. We chose a smaller number to
further challenge the parametrization. The result is a dimensionality reduction
of two orders of magnitude, from $4096=64\times64$ to $30$.

The network is trained using a popular gradient update scheme called
Adam, with $\beta_1=0.5$, $\beta_2=0.999$ (see~\citep{kingma2014adam}). We
use a step size of $10^{-4}$, batch size of $32$, and clipping interval
$[-0.01,0.01]$. We perform $5$ discriminator iterations per generator
iteration. In our experiments, convergence was achieved in around \num{20000}
generator iterations. The total training time was around 30 minutes using an
Nvidia GeForce GTX Titan X GPU. During deployment, the model can generate
realizations of $64\times 64$ size at the rate of about \num{5500} realizations per second.

In~\Cref{fig:uncond_samples,fig:cond_samples} we show unconditional and
conditional realizations generated by our trained models, and realizations
generated by snesim (reference). We also show realizations generated with
principal component analysis (PCA), retaining 75\% of the energy. We see that
our model clearly reproduces the visual patterns present in the prior
realizations.
In~\Cref{fig:permhist} we show histograms of the permeability at 10 randomly
selected locations, based on sets of 5000 fresh realizations generated by
snesim (i.e. not from the prior set) and by WGAN.
We find that our model generates values that are very close to either $0$ or
$1$, and almost no value in between (no thresholding has been performed at this
stage, only shifting and scaling to move the $\tanh$ interval $[-1,1]$ to
$[0,1]$, i.e. $(x+1)/2$). The histograms are remarkably close.

\subsection{Assessment in uncertainty quantification}


\begin{figure}
  \centering
  \begin{subfigure}{\textwidth}
    \includegraphics[width=\textwidth]{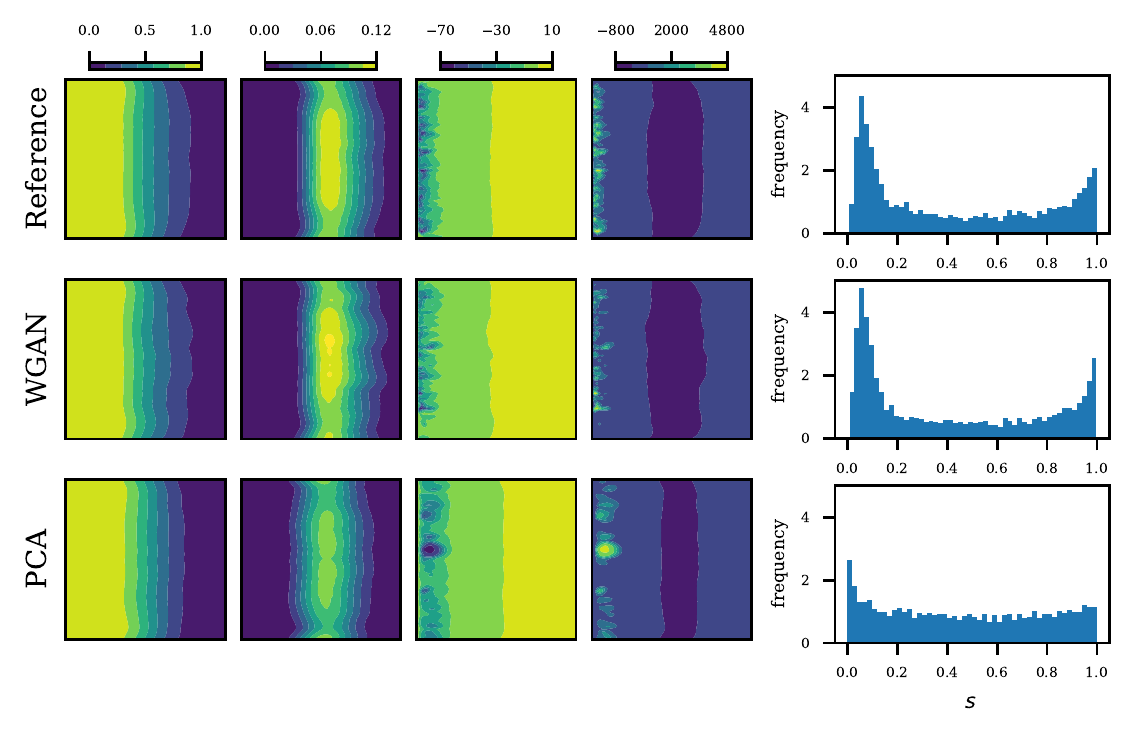}
    \vspace{-2em}\caption{Uniform flow, unconditional realizations}\vspace{2em}
  \end{subfigure}
  \begin{subfigure}{\textwidth}
    \includegraphics[width=\textwidth]{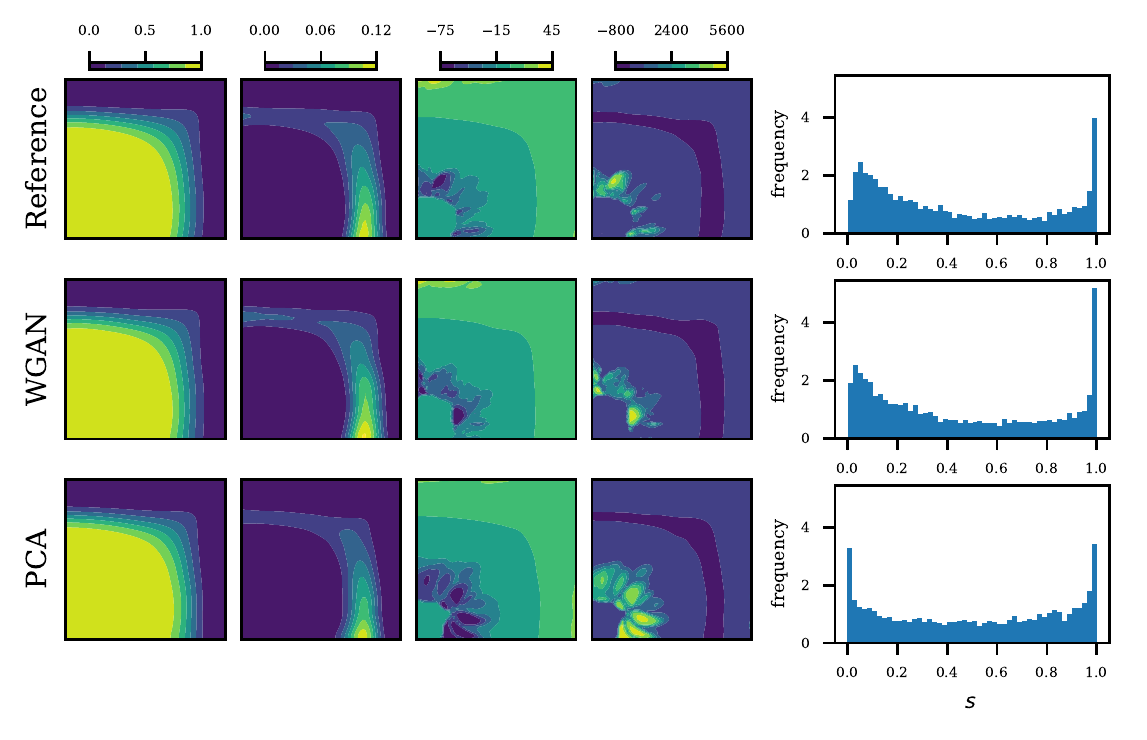}
    \vspace{-2em}\caption{Quarter five, unconditional realizations}\vspace{2em}
  \end{subfigure}
  \caption{Saturation statistics at $t=0.5\;\mathrm{PVI}$ for
    \emph{unconditional realizations}. From left to right: mean, variance,
    skewness and kurtosis of the saturation map, and lastly the saturation
    histogram at a given point. The point corresponds to the maximum variance in the
    reference.}
  \label{fig:uncond_sat}
\end{figure}

\begin{figure}
  \centering
  \begin{subfigure}{\textwidth}
    \includegraphics[width=\textwidth]{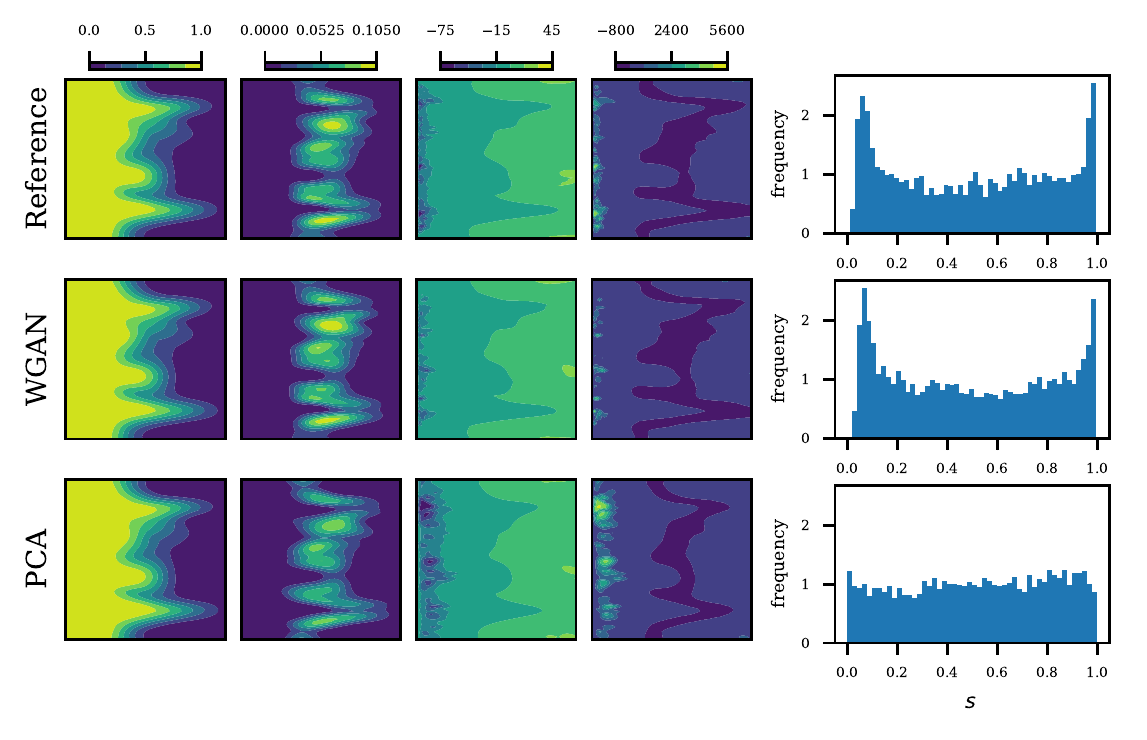}
    \vspace{-2em}\caption{Uniform flow, conditional realizations}\vspace{2em}
  \end{subfigure}
  \begin{subfigure}{\textwidth}
    \includegraphics[width=\textwidth]{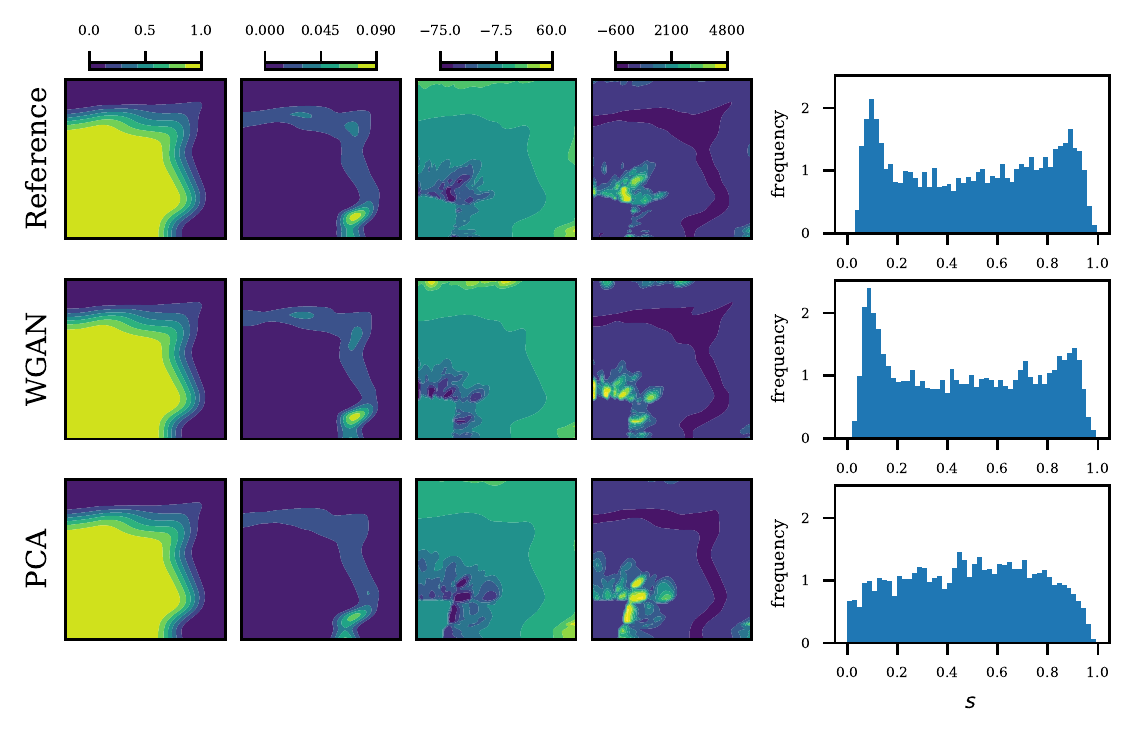}
    \vspace{-2em}\caption{Quarter five, conditional realizations}\vspace{2em}
  \end{subfigure}
  \caption{Saturation statistics at $t=0.5\;\mathrm{PVI}$ for
    \emph{conditional realizations}. From left to right: mean, variance,
    skewness and kurtosis of the saturation map, and lastly the saturation
    histogram at a given point. The point corresponds to the maximum variance in the
    reference.}
  \label{fig:cond_sat}
\end{figure}


\begin{figure}
  \centering
  \begin{subfigure}{\textwidth}
    \includegraphics[width=\textwidth]{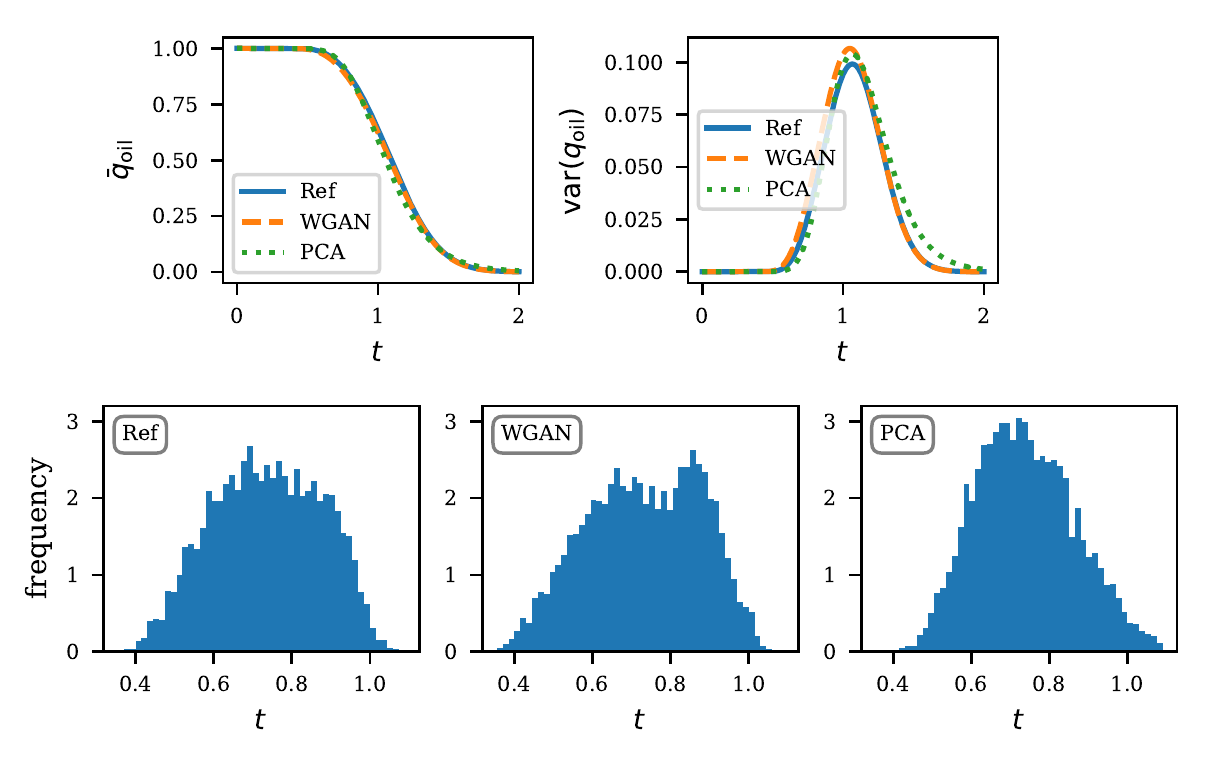}
    \vspace{-2em}\caption{Uniform flow, unconditional realizations}\vspace{2em}
  \end{subfigure}
  \begin{subfigure}{\textwidth}
    \includegraphics[width=\textwidth]{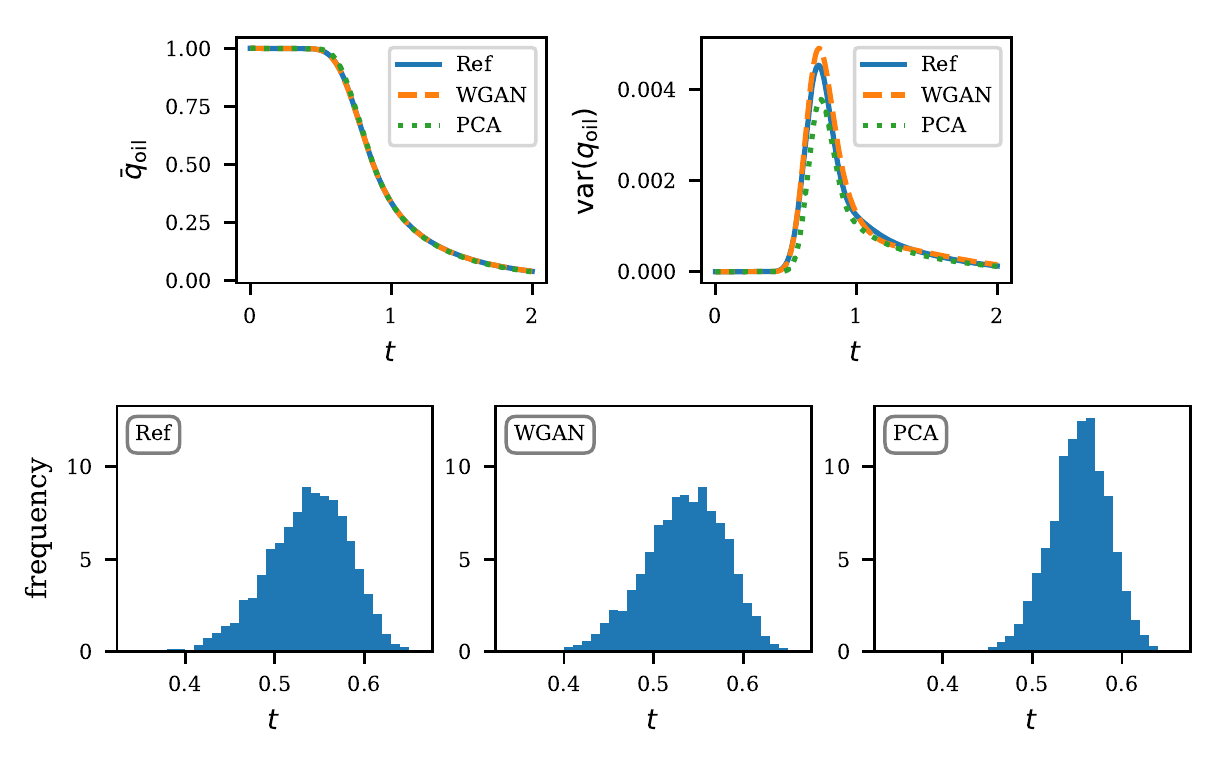}
    \vspace{-2em}\caption{Quarter five, unconditional realizations}\vspace{2em}
  \end{subfigure}
  \caption{Production statistics for \emph{unconditional realizations}. The
    top of each subfigure shows the mean and variance of the production curve.
    The bottom shows the histogram of the water breakthrough time. Times
    are expressed in pore volume injected.}
  \label{fig:uncond_prod}
\end{figure}

\begin{figure}
  \centering
  \begin{subfigure}{\textwidth}
    \includegraphics[width=\textwidth]{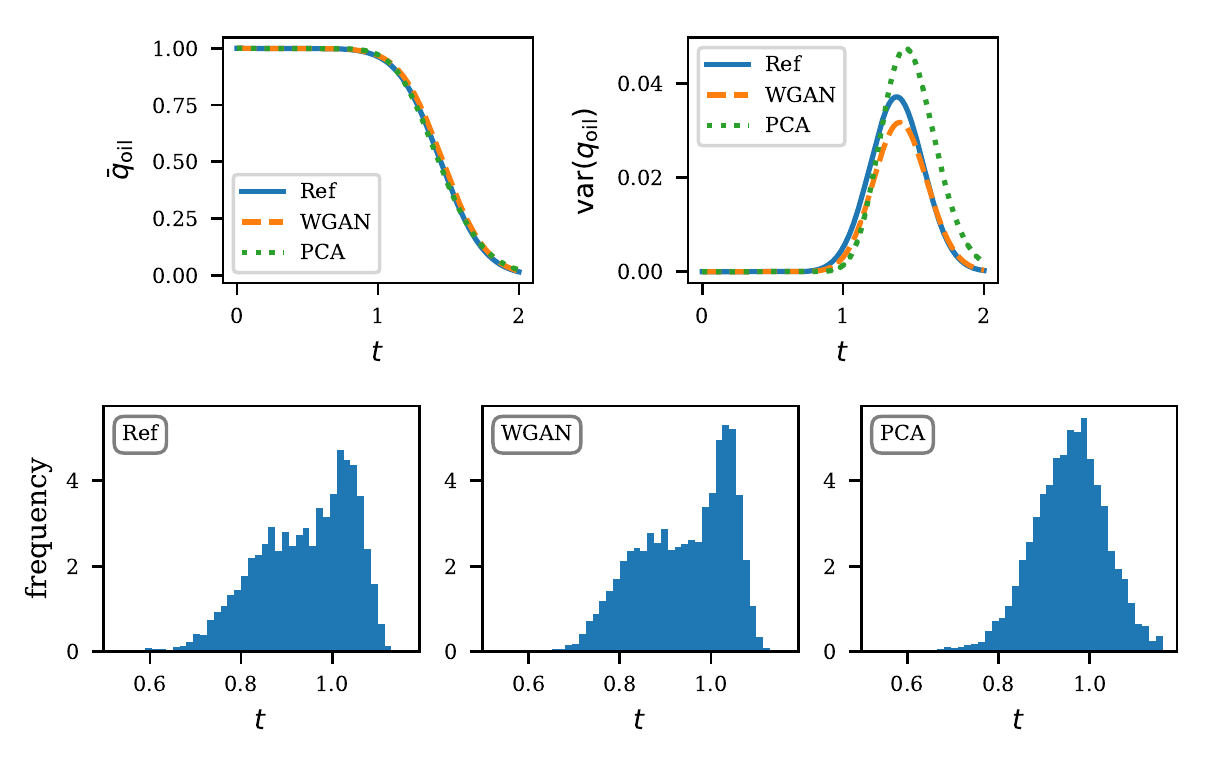}
    \vspace{-2em}\caption{Uniform flow, conditional realizations}\vspace{2em}
  \end{subfigure}
  \begin{subfigure}{\textwidth}
    \includegraphics[width=\textwidth]{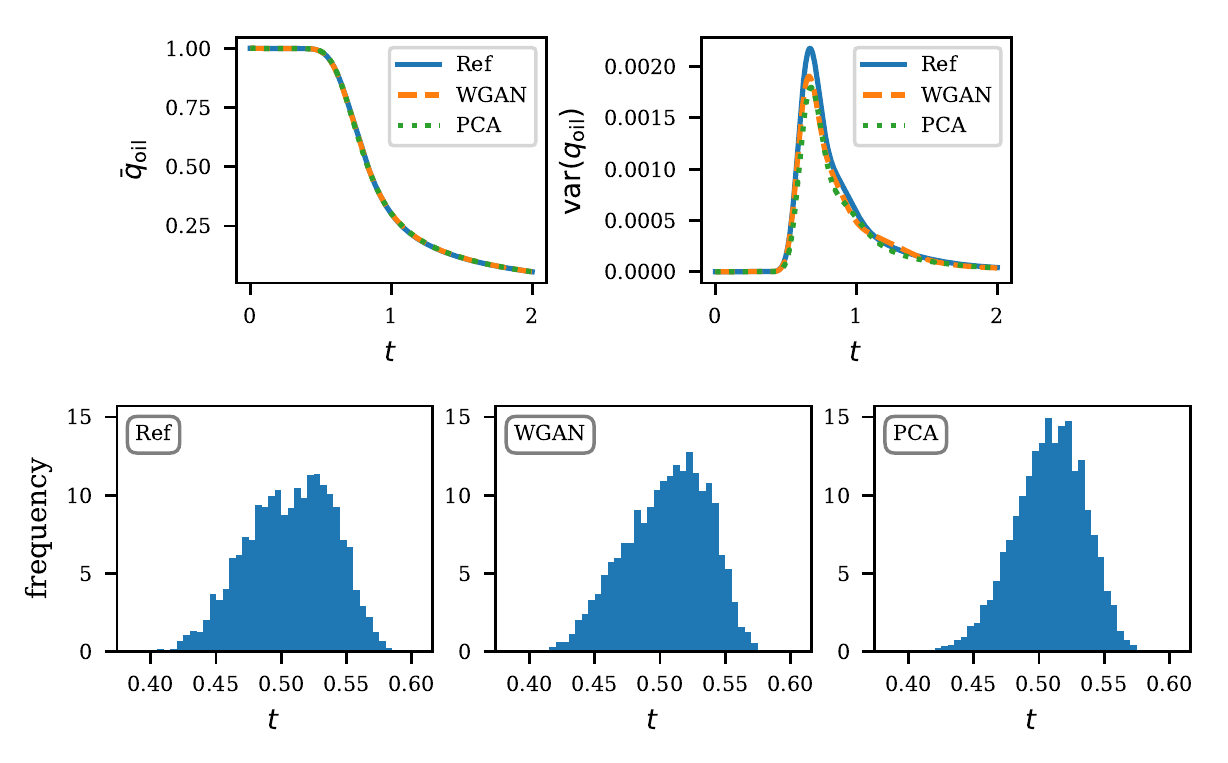}
    \vspace{-2em}\caption{Quarter five, conditional realizations}\vspace{2em}
  \end{subfigure}
  \caption{Production statistics for \emph{conditional realizations}. The top
    half of each subfigure shows the mean and variance of the production curve.
    The bottom show the histogram of the water breakthrough time. Times are
    expressed in pore volume injected.}
  \label{fig:cond_prod}
\end{figure}

In this section, we perform uncertainty quantification and estimate several
flow statistics of interest. We borrow test cases from~\citep{ma2011kernel},
where the authors parametrize the same type of permeability using Kernel PCA.
Thus, we refer the interested reader to such work for results using Kernel
PCA. Note that here we use a larger grid ($64\times64$ vs $45\times45$) and
provide an additional flow test case.

We propagate 5000 realizations of the permeability field in 2D single-phase subsurface flow. 
We consider injection of water for the purpose of displacing oil inside a reservoir (water and oil in this case have the same fluid properties since we consider single-phase flow).
The system of equations for this problem is

\begin{align}
  -\nabla\cdot(a\nabla p) &= q \label{eq:pressure} \\
  \varphi\frac{\partial s}{\partial t} + \nabla\cdot(sv) &= q_w \label{eq:saturation}
\end{align}
where $p$ is the fluid pressure, $q=q_w+q_o$ denotes (total) fluid sources, $q_w$ and $q_o$ are the water and oil sources, respectively, $a$ is the permeability, $\varphi$ is the porosity, $s$ is the saturation of water, and $v$ is the Darcy velocity. 

Our simulation domain is the unit square with $64\times64$ discretization grid. The reservoir initially contains only oil, i.e. $s(\mathbf x,t=0)=0$, and we simulate from $t=0$ until $t=0.4$. We assume an uniform porosity of $\varphi=0.2$. 
We consider two test cases:

{\setlist[description]{font=\normalfont\itshape\space}
\begin{description}
\item[Uniform flow:] We impose uniformly distributed inflow and outflow
  conditions on the left and right sides of the unit square, respectively, and
  no-flow boundary conditions on the remaining top and bottom sides. The total
  absolute injection/production rate is $1$. For the unit square, this means
  $v\cdot\hat{n}=-1$ and $v\cdot\hat{n}=1$ on the left and right sides,
  respectively, where $\hat{n}$ denotes the outward-pointing unit normal to the
  boundary.
\item[Quarter-five spot:] We impose injection and production points at $(0,0)$
  and $(1,1)$ of the unit square, respectively. No-flow boundary conditions are
  imposed on all sides of the square. The absolute injection/production rate is $1$, i.e.
  $q(0,0)=1$ and $q(1,1)=-1$.
\end{description}
}

The propagation is done on sets of realizations generated by WGAN and by snesim
for comparison. Note that these are fresh realizations not used to train the
WGAN models. We also show results using PCA for additional comparison.

Statistics of the saturation map based on 5000 realizations are summarized
in~\Cref{fig:uncond_sat,fig:cond_sat}. We plot the saturation at time $t=0.1$,
which corresponds to $0.5$ pore volume injected (PVI).
From left to right, we plot the mean, variance, skewness and kurtosis of the saturation
map. We see that the statistics from realizations generated by WGAN correspond
very well with the statistics from realizations generated by snesim
(reference). We also see that the PCA parametrization performs very well in the mean and variance,
however the discrepancies increase as we move to higher order moments. The discrepancy
becomes clearer by plotting the histogram of the 5000 saturations at a fixed point in
the domain, shown on the last columns of~\Cref{fig:uncond_sat,fig:cond_sat}. We choose the point in the domain where reference saturation had the highest variance. We see that the histograms by WGAN match the reference
remarkably well even in multimodal cases. The reader may compare our
results with~\citep{ma2011kernel}. The results suggest that the generator
effectively learned to replicate the data generating process.

Statistics of the production curve are summarized
in~\Cref{fig:uncond_prod,fig:cond_prod}. On the top half of each subfigure, we
show the mean and variance of the production curve based on $5000$ realizations.
These can in general be approximated well enough by using only the PCA parametrization. We find that
the performance of our models are also comparable for this task.
To further contrast the ability to preserve higher order statistics, we plot the
histogram of the water breakthrough time, for which an accurate estimation is of
importance in practice. Here we define the water breakthrough time as the time
that water level reaches 1\% of production. Results are shown on the bottom half
of each subfigure in~\Cref{fig:uncond_prod,fig:cond_prod}. In all cases, we find
a very good agreement between WGAN and reference. Unlike PCA, the
responses predicted by WGAN do not have a tendency to be normally
distributed (see e.g.~\Cref{fig:cond_prod}a).

\subsection{Assessment in parameter estimation} \label{sec:assimilation}

\begin{figure}
  \centering
  \includegraphics[width=\textwidth]{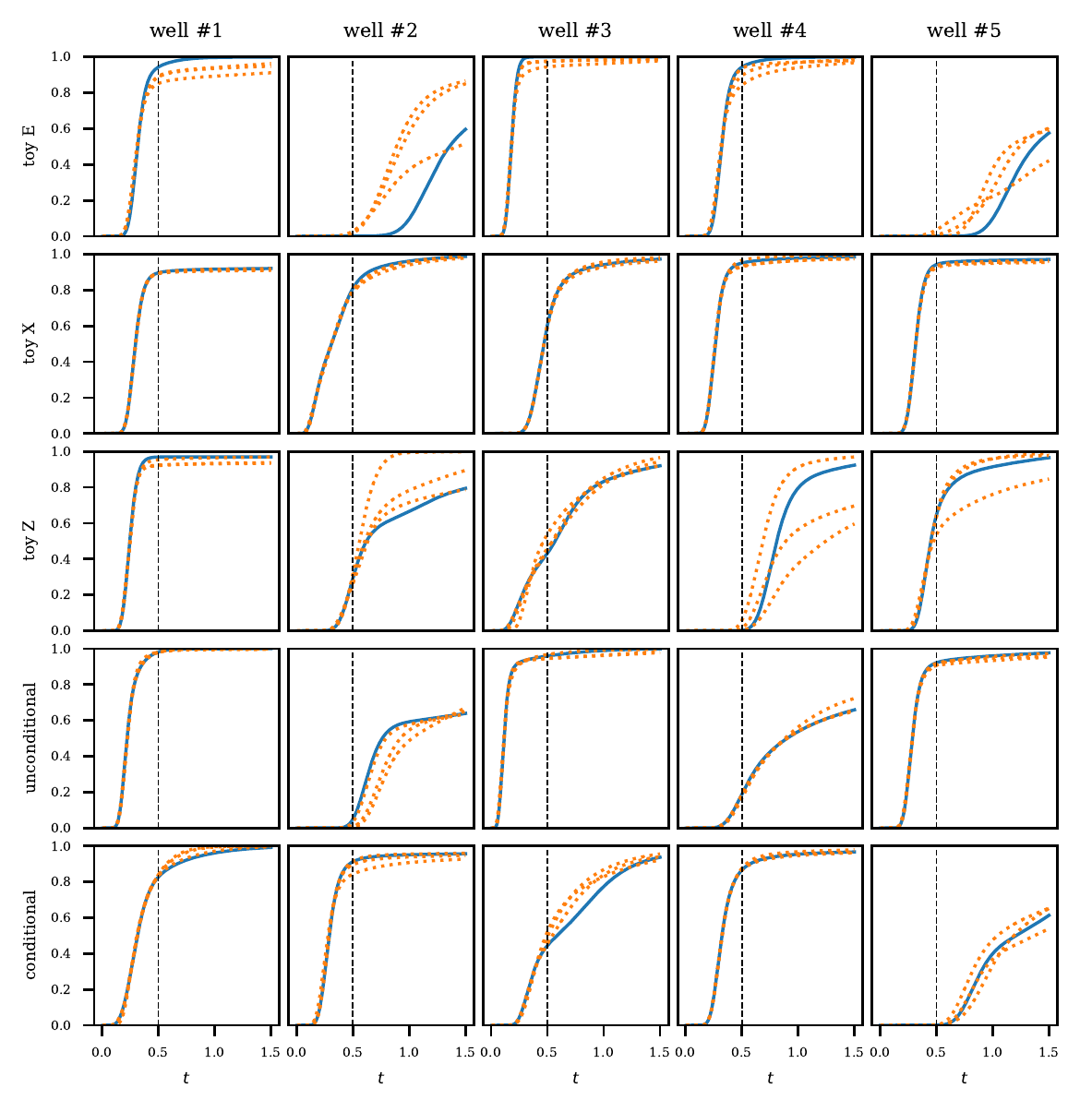}
  \caption{History matching results. Water level curves from the production
    wells in different test cases. Blue solid lines denote the target responses.
    Orange dotted lines are three matching solutions found in the inversion. The
    black vertical dashed line in each plot marks the end of the observed
    period. Times are expressed in pore volume injected.}
    \label{fig:histmatch_qout}
\end{figure}

\begin{figure}
  \centering
  \includegraphics[width=\textwidth]{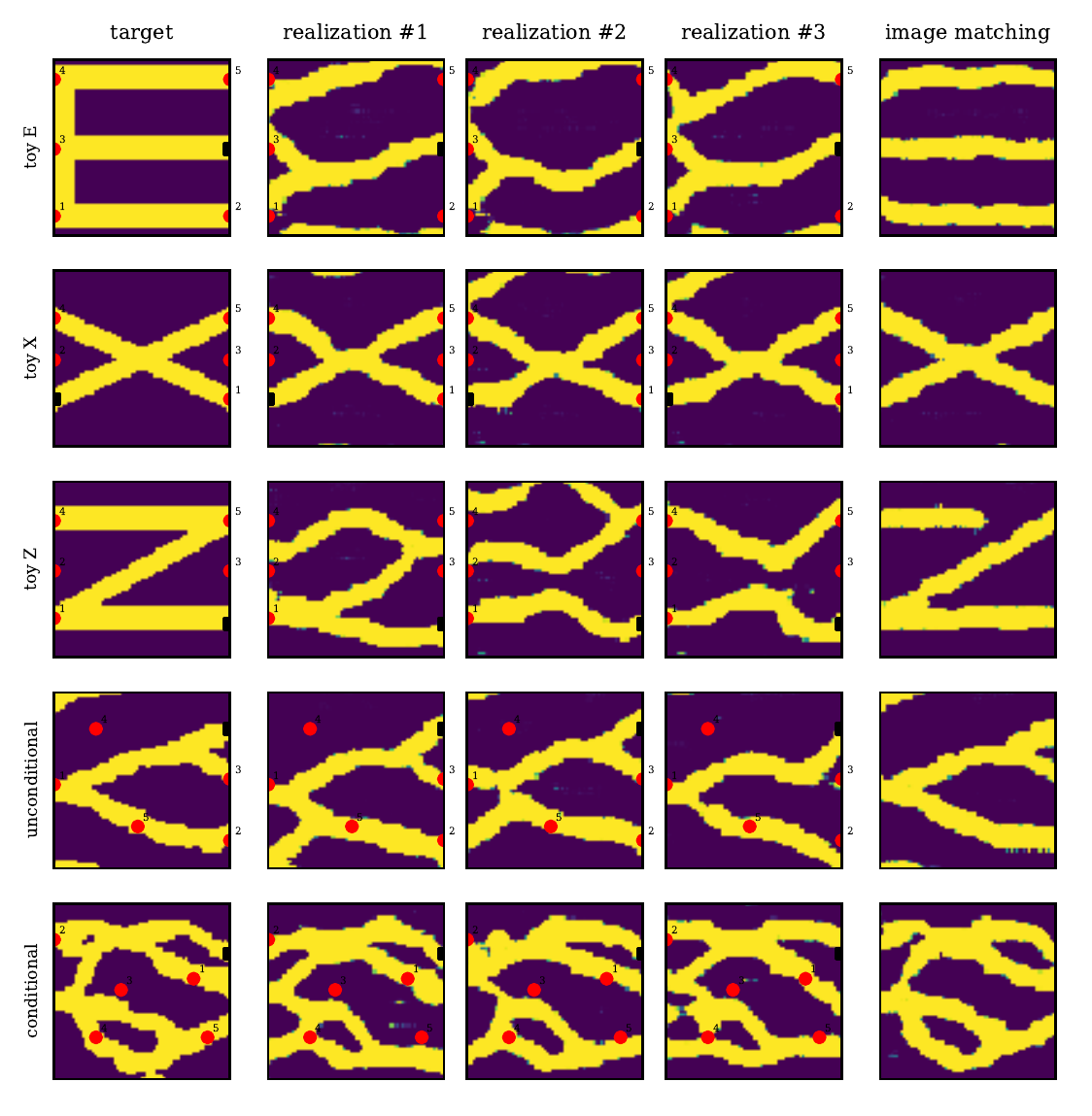}
  \caption{History matching results. We experiment with three toy images as well as unconditional and conditional snesim realizations. Each case contains one injection well (black square) and five production wells (red circles). We show three solutions that match the observed production period (see~\Cref{fig:histmatch_qout}). The last column contains image matching solutions.}
  \label{fig:histmatch}
\end{figure}

We now assess our models for parameter estimation where we reconstruct the subsurface permeability based on historical data of the oil production stage, also known as history matching. Following the general problem setting from before, we aim to find realizations of the permeability that match the production curves observed from wells.

\subsubsection{Inversion using natural evolution strategies}
Let $ \dd=\mathcal{M}(\mathbf a)$ where $\mathcal M$ is the forward map,
mapping from permeability $\mathbf a$ to the output
$\mathbf d$ being monitored (in our case, the water level curve at the
production wells). Given observations $\mathbf d_{\mathrm{obs}}$ and assuming i.i.d.
Gaussian measurement noise (we use $\sigma=0.01$), and prior $\mathbf
z\sim\mathcal{N}(\mathbf 0,\mathbf I)$, the objective function to be maximized is

\begin{align}
	f(\zz) & = - \frac{1}{\sigma^2}(\dd - \dd_{\mathrm{obs}})^T (\dd - \dd_{\mathrm{obs}}) -  \zz^T \zz                                                                  \\
	       & = - \frac{1}{\sigma^2}(\mathcal M(G(\mathbf z))   -\mathbf d_{\mathrm{obs}})^T (\mathcal M(G(\mathbf z))   -\mathbf d_{\mathrm{obs}}) - \mathbf z^T \mathbf z 
\end{align}

To maximize this function, we use natural evolution strategies (NES)~\citep{wierstra2008natural,wierstra2014natural}, a black-box optimization method suitable for the low-dimensional parametrization achieved. Another reasonable alternative is to use gradient-based methods exploiting the differentiability of our generator. This would require adjoint procedures to get the gradient of the forward map $\mathcal M$. We adopted NES due to its generality and easy implementation that does not involve the gradient of $f$ (nor $\mathcal M$). NES maximizes $f$ by maximizing an average of $f$ instead,
$J(\phi)\coloneqq \mathbb{E}_{{\pi(\mathbf z|\phi)}}{f(\mathbf z)}$,
where $\pi(\zz|\phi)$ is some distribution parametrized by $\phi$ (e.g.
we used the family of Gaussian distributions, in which case $\phi$ involves
the mean and covariance matrix). This is based on the observation that
$\max_\phi\mathbb{E}_{{\pi(\zz|\phi)}}{f(\mathbf z)} \leq \max_{\zz} f(\zz)$.
%

Optimizing the expectation $\mathbb{E}_{{\pi(\zz|\phi)}}{f(\mathbf z)}$ (instead of optimizing $f$ directly) has the advantage of not requiring the gradient of $f$ (and therefore of the simulator) since

$$\nabla_{\phi} J(\phi) = \mathbb E_{{\pi(\zz|\phi)}}{f(\mathbf z)\nabla_\phi{\log{\pi(\mathbf z|\phi}}})$$
We can approximate this as

$$\nabla_{\phi} J(\phi) \approx \frac{1}{N}\sum^N_{k=1}{ f(\mathbf z_k)\nabla_\phi\log{\pi(\mathbf z_k|\phi)} }$$
by drawing realizations $\mathbf z_1,\cdots,\mathbf z_N \sim \pi(\mathbf z|\phi)$.
Optimization proceeds by simple gradient ascent, $\phi \gets \phi +
\eta\nabla_\phi J(\phi)$ where $\eta$ is a step size. Note that we optimize
the parameter of the search distribution $\phi$, rather than $\mathbf z$. As
the optimization converges, the search distribution collapses to an optimal
value of $\mathbf z$. In our implementation, we actually use an improved version
of NES which uses the Fisher matrix and natural coordinates, as detailed
in~\citep{wierstra2014natural}.

\subsubsection{History matching}
We consider five target images of the permeability: 
one unconditional realization and one conditional realization (both using snesim), and three hand-crafted images (see first column in~\Cref{fig:histmatch}). The latter were specifically designed to test the limits of the parametrization.
For the conditional realization, we use the generator trained on conditional realizations. For the remaining cases, we use the unconditional generator. Note that this poses a difficulty on the reconstruction of the hand-crafted toy problems as these are not plausible realizations with respect to the dataset used to train the generator.

In each test case, we set one injection well with fixed flow rate of $1$, and five production wells with flow rate of $-0.2$ (locations marked on each image, see~\Cref{fig:histmatch}). Our only observed data are the water level curves at the production wells from $t=0$ to $t=0.5\;\mathrm{PVI}$, induced by the target permeabilities. We do not include knowledge of the permeability at the ``drilled'' wells (as normally done in real applications) in the parameter estimation.
For the experiments, we scaled the log-permeability values of $0$ and $1$ to $0$ and $5$, emulating a shale and sand scenario. We have done this in part to allow for a less underdetermined system 
(i.e. so that different permeability patterns produce more distinctive flow patterns).

Results for history matching are shown in~\Cref{fig:histmatch,fig:histmatch_qout}. For each test case, we find three solutions of the inversion problem using different seeds (initial guess). For the conditional and unconditional realizations, we obtain virtually perfect match of the observed period (\Cref{fig:histmatch_qout}). Beyond the observed period, the responses naturally diverge. As is expected, the matching is more difficult for some of the toy problems, in particular toy problem E and toy problem Z. Toy problem X, however, does particularly well.

In~\Cref{fig:histmatch} we show the reconstructed permeability images for each test case. We also
show, in the last column, image matching solutions (i.e. we invert conditioning on the whole image using NES).
For the conditional and unconditional cases, we see a good visual correspondence between target and solution realizations in the history matching. We also find good image matching solutions, verifying that the target image is in the solution space of the generator and therefore the history matching can be further improved by supplying more information (e.g. permeability values at wells, longer historical data, etc.). This applies to toy problem X as well, where the target seems to be plausible (high probability under the generator's distribution).
As expected, the reconstruction is more difficult for toy problems E and Z, where the target images seem to have a low probability as suggested by the image matching solutions. For these cases, history matching will cease to improve after certain point. Note that this is not a failure of the parametrization method -- after all, the generator should be trained using prior realizations that inform the patterns and variability of the target.
That is, the parametrization must be done using samples deemed representative of the geology under study.

\subsection{Honoring point conditioning}

\begin{table}
  \begin{subtable}{\textwidth}
  \centering
  \begin{tabular}{r| c c c c}
           & $j=12$ & $j=25$ & $j=38$ & $j=51$ \\
    \hline
    $i=12$ & $0.38$ & $0.48$ & $1.78$ & $1.82$ \\
    $i=25$ & $0.06$ & $0.34$ & $0.54$ & $0.02$ \\
    $i=38$ & $4.46$ & $ 1.3$ & $ 3.0$ & $ 0.8$ \\
    $i=51$ & $2.06$ & $1.16$ & $ 1.2$ & $0.26$ \\
  \end{tabular}
  \caption{Percentage of mismatches at each conditioning point.}
  \label{table:misatpoint}
  \end{subtable}
  \vspace{1em}

  \begin{subtable}{\textwidth}
  \centering
  \begin{tabular}{r|c c c}
                   & one or more & two or more & three \\
    \hline
    exact          & $17.6$ & $1.82$ & $0.12$\\
    1 cell away    & $ 1.8$ & $0.02$ & $ 0.0$\\
    2 cells away   & $0.46$ & $ 0.0$ & $ 0.0$\\
    3 cells away   & $0.24$ & $ 0.0$ & $ 0.0$\\
    4 cells away   & $0.08$ & $ 0.0$ & $ 0.0$
    \end{tabular}
  \caption{Percentage of realizations with mismatches.}
  \label{table:misreal}
  \end{subtable}
  \caption{Performance in honoring point conditioning.}
  \label{table:mismatch}
\end{table}

\begin{figure}
  \centering
  \begin{subfigure}{\textwidth}
    \centering
    \includegraphics[scale=0.65]{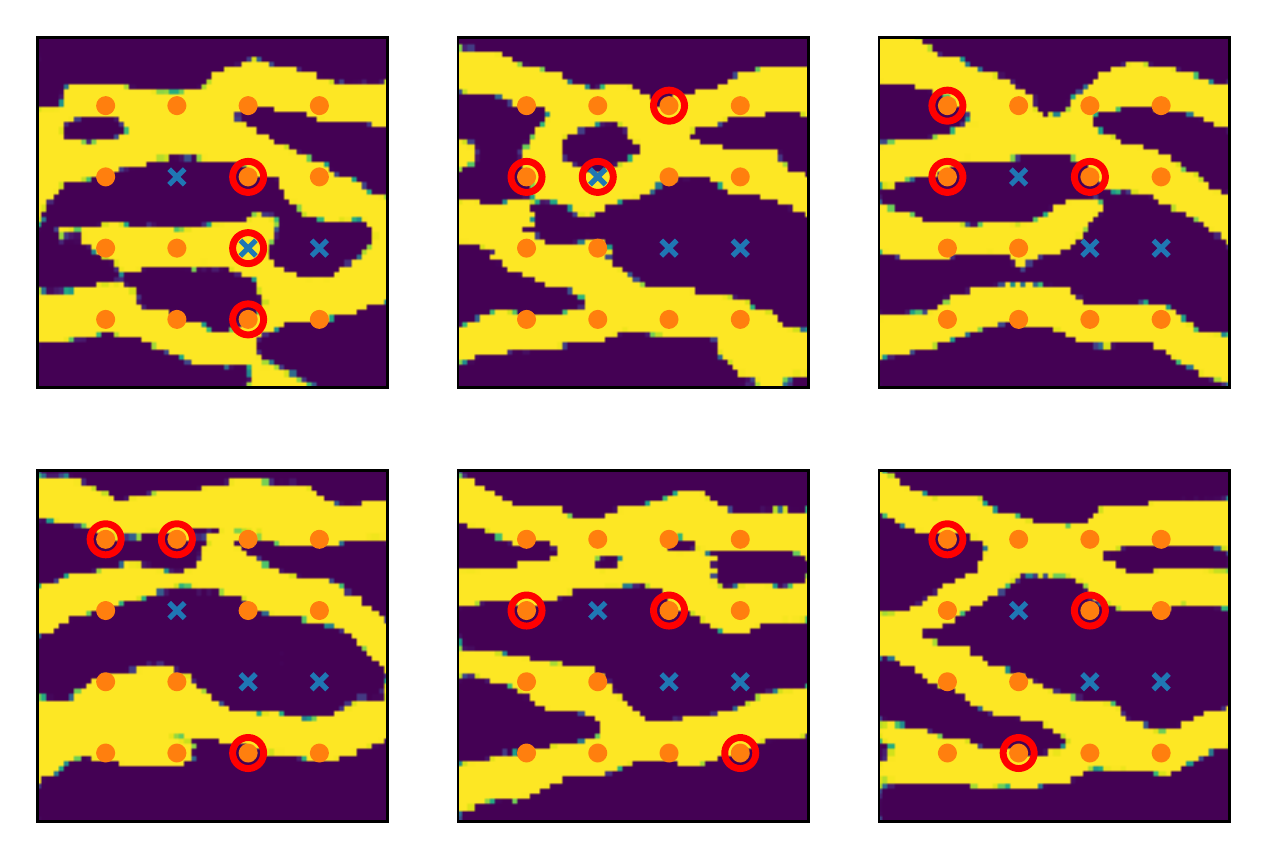}
    \vspace{-1pt}\caption{Realizations containing $3$ mismatches.}\vspace{10pt}
    \label{fig:3mismatch}
  \end{subfigure}
  \begin{subfigure}{\textwidth}
    \centering
    \includegraphics[scale=0.65]{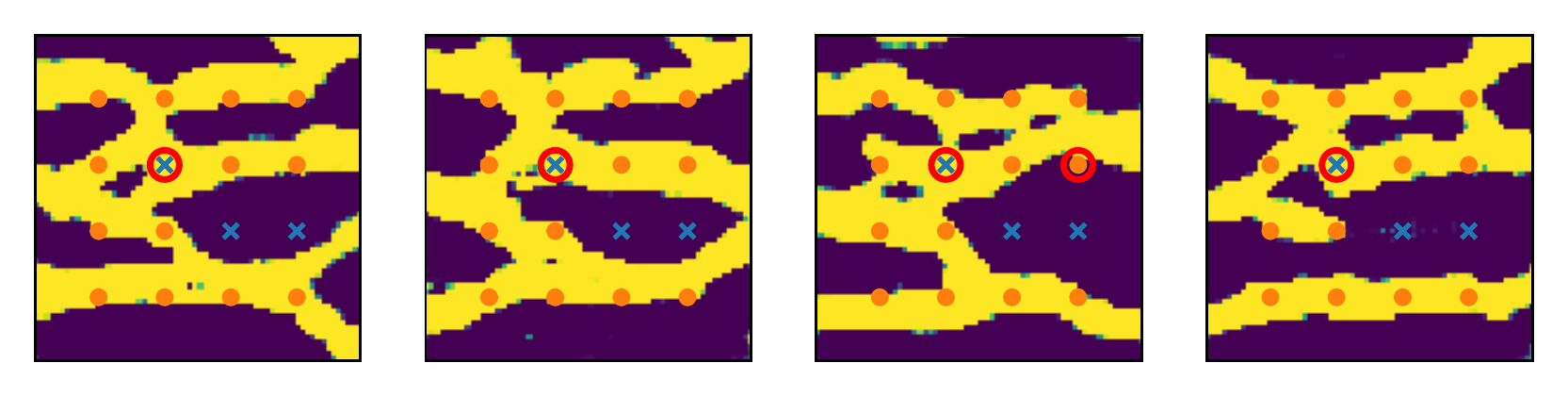}
    \vspace{-1pt}\caption{Realizations with large misplacement ($4$ cells away).}\vspace{1pt}
    \label{fig:bigmismatch}
  \end{subfigure}
  \caption{Realizations where conditioning failed. Orange dots indicate
    points conditioned to low permeability ($0$) and blue crosses indicate
    points conditioned to high permeability ($1$). Mismatches are circled in
    red.}
\end{figure}

We assess the ability of the generator trained on conditional
realizations to reproduce the point conditioning. We analyze $5000$ realizations
and report in~\Cref{table:misatpoint} the percentage of mismatches at each of
the $16$ conditioning points. We find that mismatches do occur at frequencies of
less than $5\%$ at each conditioning point. Next, we count the overall number of realizations with at least
$1$ mismatch, at least $2$ mismatches, and $3$ mismatches (there were no
realizations with more than $3$ mismatches). The result is reported
in~\Cref{table:misreal}. The first row shows the percentage of realizations that contain mismatches. 
We see that $82.4\%$ of realizations honor all conditioning points.
Of the sizable $17.6\%$ of realizations that do contain mismatches, most have only $1$ mismatch.
In particular, we find only $6$ (0.12\%) realizations containing three mismatches,
shown in~\Cref{fig:3mismatch}. From the figure, we notice that most mismatches
were misplaced by a few cells. 
We find that this is generally the case:
In~\Cref{table:misreal}, we report the percentage of realizations that contain mismatches with misplacement of $1$, $2$, $3$, and $4$ cells (there were no larger misplacement). We find that if
we allow for a tolerance of $1$ cell distance, the percentage of wrong realizations
drops to less than $2\%$. Specifically, $98.2\%$ of realizations honor all
conditioning points within a $1$ cell distance, and $82.4\%$ do so exactly. This could
explain the yet good results in flow experiments. Finally, we
show in~\Cref{fig:bigmismatch} the only $4$ ($0.08\%$) realizations containing large
misplacement of $4$ cells.

Note that mismatches do not occur using PCA parametrization (assuming an exact method for the eigendecomposition is used) as it is derived to explicitly preserve the spatial covariances. The presence of mismatches in our method reflects the approach that we take to parametrization:
We formulate the parametrization by addressing the data generating process rather than the spatial statistics of the data, resulting in a parametrization that extrapolates to new realizations that, except for a few pixels/cells, are otherwise indistinguishable from data. In view of the good results from our flow experiments, the importance of honoring point conditioning precisely to the cell level could be argued. On the other hand, we also acknowledge that conditioning points are normally scarce and obtained from expensive measurements, so it is desirable that these be well honored in the parametrization.

\section{Discussion and practical details} \label{sec:practical}
\subsection{Practical advantages of WGAN} \label{sec:importance}

\begin{figure}
  \centering
  \includegraphics[width=\textwidth]{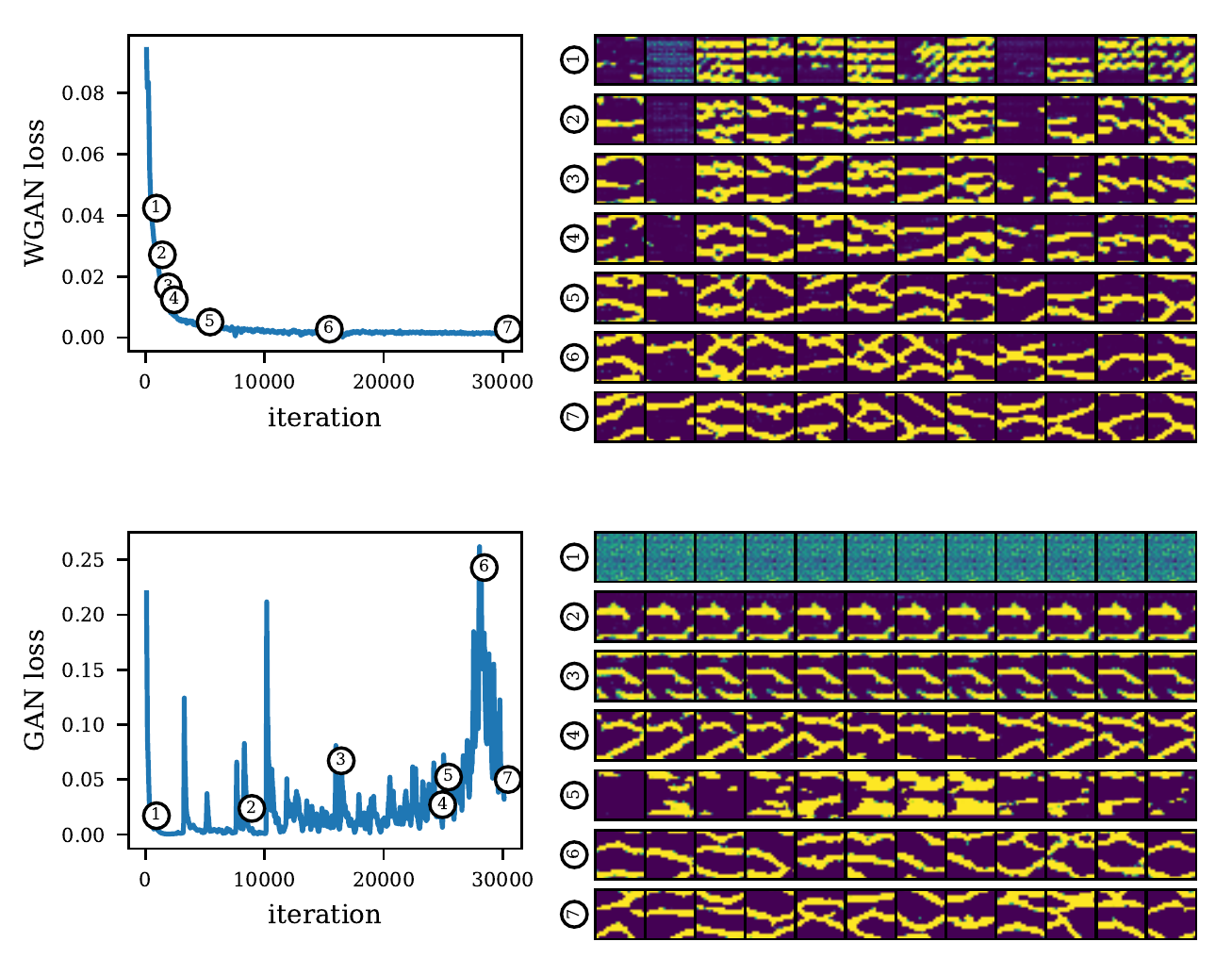}
  \caption{Convergence curves of a WGAN model (top) and a standard GAN model
    (bottom). On the right, we show realizations along the training of the
    corresponding models. We see that GAN loss is uninformative regarding sample
    quality. Note that the losses are not comparable between models since the
    formulations are different.}
  \label{fig:convergence}
\end{figure}

\begin{figure}
  \centering
  \includegraphics[width=\textwidth]{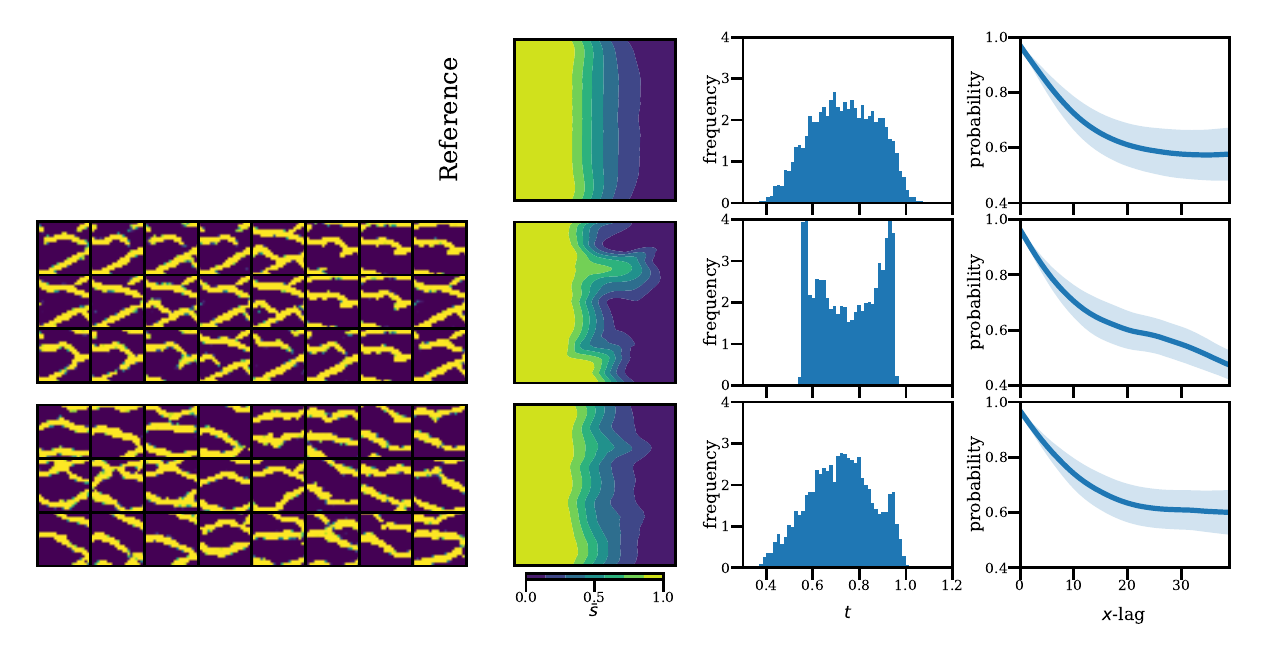}
  \caption{Examples of missing modes in standard GAN. Second and third rows show
    realizations generated by collapsing GAN models (left) and their responses
    (right). First row shows the reference solutions.
    The standard GAN was trained using the same generator architecture, but a
    $\times 4$ larger discriminator than the one used in WGAN. We did not manage to find
    convergence with smaller discriminator sizes.
  }
  \label{fig:collapse}
\end{figure}

An issue with the standard formulation of GAN is the lack of a convergence
curve or loss function that is informative about the sample quality.
We illustrate this in~\Cref{fig:convergence} where we show the convergence
curve of our trained WGAN model, and a convergence
curve of a GAN using the standard formulation. We also show realizations
generated by the models along the training process.
The curve of WGAN follows the ideal behavior that is expected in an optimization
process, whereas the curve of standard GAN is erratic and shows no correlation
with the quality of the generated samples. We can also see another well-known
issue of standard GAN which is the tendency to mode collapse, i.e. a lack of
sample diversity, manifested as the repetition of only one or few image modes. We
see that the standard GAN generator jumps from one mode solution to another.
Note that in some cases, however, mode collapse is more subtle and not easily
detectable. This is very problematic to our application since it can lead to
biases in uncertainty quantification and unsuccessful history matching due to
the absence of some modes in the generator.

Given the lack of an informative convergence metric in standard GAN, the
training process would involve a human judge serving as the actual loss
function to track the visual quality along the training (in practice, weights are
saved at several checkpoints and assessed after the fact).
On top of this, the human operator would need to look at multiple realizations at once in an attempt to
detect mode collapse.
Clearly, this subjective process is error prone, not to mention labor intensive.
In~\Cref{fig:collapse} we show two standard GAN models and their flow responses
in the unconditional uniform flow test case, based again on $5000$ realizations.
On the top row, we show again the reference results (mean saturation and water
breakthrough time) for comparison. We also compute the two-point probability
function~\citep{torquato1982} of the generated realizations (last column; we
show the mean and one standard deviation). We see that in some cases, mode
collapse is very evident and the model can be quickly discarded (second row). In
other cases (last row), mode collapse is harder to detect and can lead to
misleading predictions.
We also see that the two-point probability function is not sufficient to detect
mode collapse as this function does not measure the overall sample diversity.

The Wasserstein formulation solves these issues by allowing the generator to minimize the Wasserstein distance between its distribution and the target distribution (\Cref{eq:wasserstein}), therefore reducing mode collapse. Moreover, the Wasserstein distance is readily available in-training and can be used to assess convergence. Therefore, the Wasserstein formulation is better suited for automated applications, where robustness and a convergence criteria is essential.

\subsection{Network sizes under limited data}\label{sec:netsize}

\begin{figure}
  \centering
  \begin{subfigure}{\textwidth}
    \includegraphics[width=\textwidth]{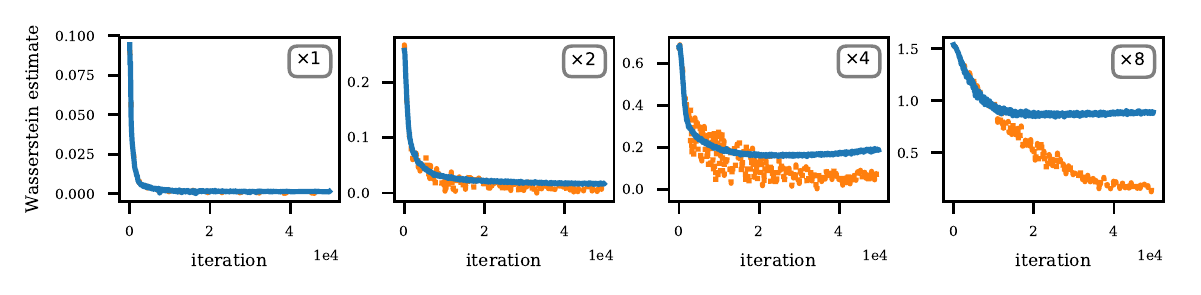}
    \caption{Convergence curve for different sizes of $D$ (and fixed $G$). Solid
      blue lines indicate the training loss, and orange dotted lines indicate
      the validation loss. Note that the losses cannot be compared since
      the Lipschitz constants are different.}
    \label{fig:compare_D}
  \end{subfigure}
  \vspace{1em}

  \begin{subfigure}{\textwidth}
    \includegraphics[width=\textwidth]{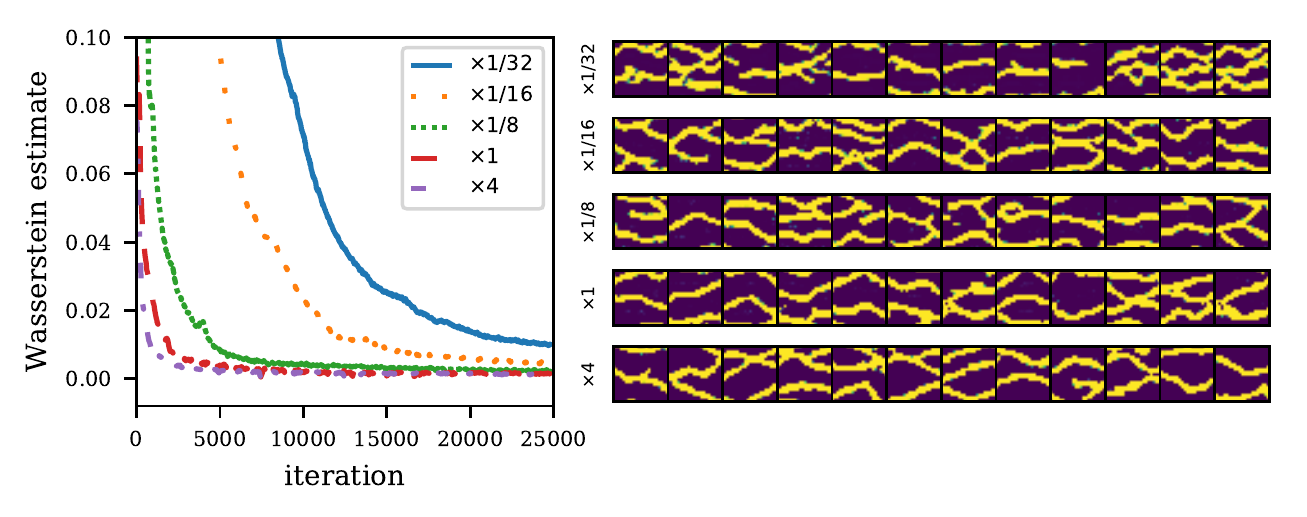}
    \caption{\emph{Left:} Convergence curve for different sizes of $G$ (and fixed $D$).
      \emph{Right:} Realizations by generators of different sizes (at \num{15000} iterations).}
    \label{fig:compare_G}
  \end{subfigure}
  \caption{Performance of models with varying network sizes.}
\end{figure}

As mentioned earlier, architecture design is largely problem-dependent and
based on domain knowledge and heavy use of heuristics. The general approach
is to start with a baseline architecture from a similar problem domain and
tune it to accommodate for the present problem. Current computer vision
applications use the pyramid architecture shown in~\Cref{fig:cnn}. These
applications benefit from very large datasets of images. In contrast, our
application uses a relatively small dataset.
Recall that the discriminator $D$ is trained using this limited dataset,
therefore addressing the possibility of overfitting is important since the accuracy of $D$ is crucial in the performance of the method. 
In particular, an overfitted $D$ creates an issue where the Wasserstein
estimate in~\Cref{eq:wasserstein} is no longer accurate, making the gradients
to the generator unreliable.
We show the effect of overfitting in~\Cref{fig:compare_D} by training models
with different discriminator sizes, and fixed generator architecture. We
train discriminators of $2$, $4$, and $8$ times the size of the discriminator
used in our previous experiments. The way we increase the model size is by
increasing the number of filters in each layer of the discriminator while
keeping everything else constant. Another possibility is to add extra layers
to the architecture.
To detect overfitting, we evaluate the Wasserstein estimate using
a separate validation set of $200$ snesim realizations. We see that for an
adequate size of the discriminator, the Wasserstein estimate as evaluated on
either training or validation set are similar. However for larger models, the
Wasserstein estimates on the training and validation sets start to wildly
diverge as the optimization progresses, suggesting that the discriminator is
overfitting and the estimates are no longer reliable. It is therefore necessary
to adjust the size of the discriminator or use regularization techniques when
data is very limited.

Regarding generator architectures, network sizes will in general be limited
by compute and time resources; on the other hand, we only need just enough
network capacity to be able to model complex structures. We illustrate this
in~\Cref{fig:compare_G} where we train generators of different sizes (like
before, we vary the number of filters in each layer) and fixed
discriminator architecture. We train generators of $\frac{1}{32}$,
$\frac{1}{16}$, $\frac{1}{8}$, and $4$ times the size of the generator used
in previous experiments. We also show realizations generated by each
generator model after \num{15000} training iterations. We see that for a very small
network model ($\frac{1}{32}$), convergence is slow (as measured by
iterations). Convergence is faster as the network size increases since it is
easier to fit a larger network. Note, however, that the iterations in larger
networks are more expensive, possibly making convergence actually slower in
terms of compute times. Moreover, a larger generator has a higher forward
evaluation cost, impacting the performance for deployment.
Therefore, after certain point it is inefficient to keep increasing the
generator size.

\subsection{GAN for multipoint geostatistical simulations}\label{sec:ch3ganmps}
\begin{figure}
  \begin{subfigure}{\textwidth}
  \centering
    \includegraphics[width=.5\textwidth]{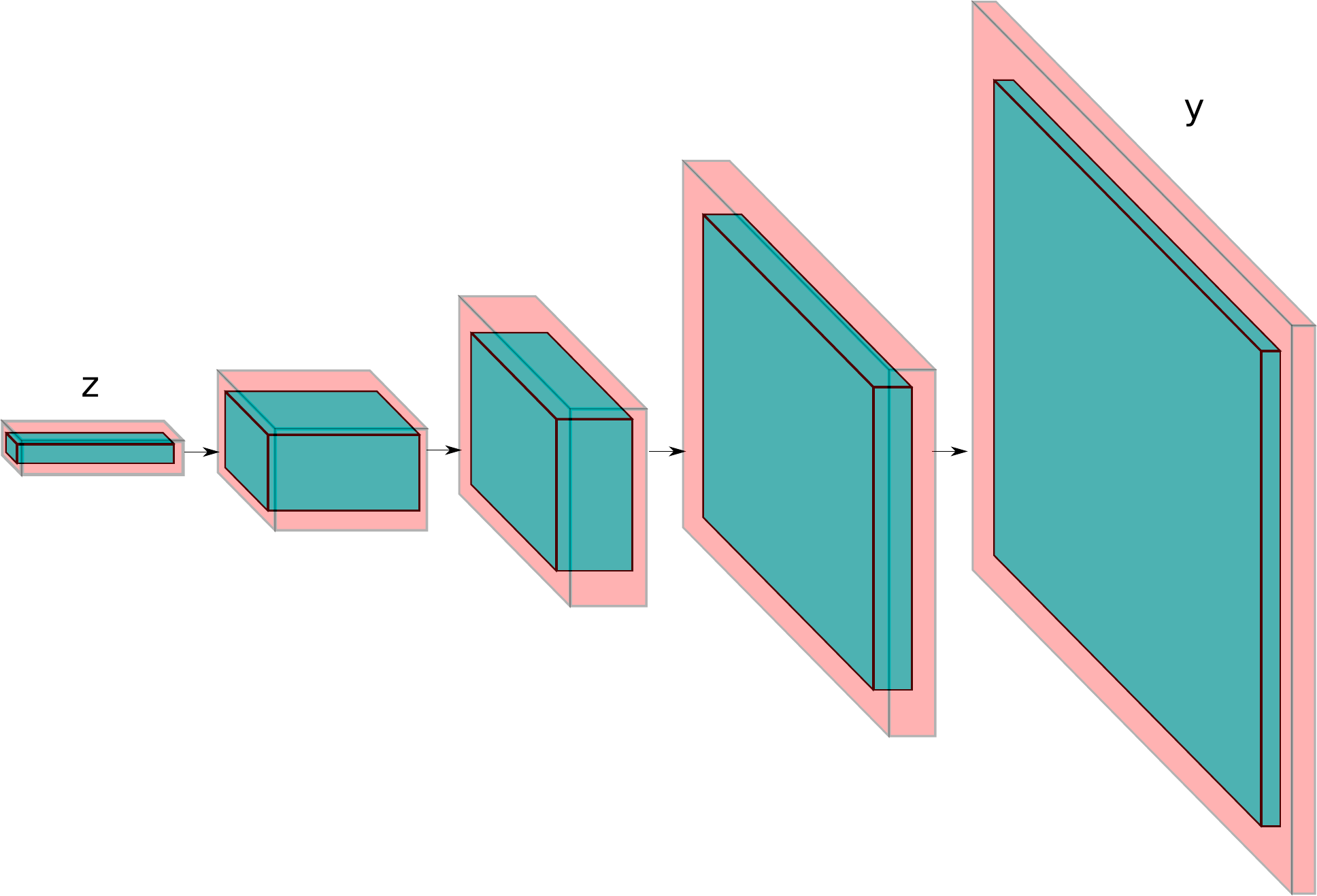}
    \caption{Illustration of artificially expanding the input array in the generator network.
      Blue blocks represent the original state shapes that a normal input array follows in the generator.
      Light red blocks represent the new state shapes of an expanded input array.
    }
    \label{fig:cnn_fake}
  \end{subfigure}
  \vspace{1em}

  \begin{subfigure}{\textwidth}
  \centering
  \begin{tikzpicture}[shorten >=1pt,->,draw=black!85, node distance=\layersep]
    \tikzstyle{every pin edge}=[<-,shorten <=1pt]
    \tikzstyle{neuron}=[circle,draw,minimum size=17pt,inner sep=0pt]
    \tikzstyle{annot} = [text width=4em, text centered]

    \foreach \name / \y in {1,...,4}
    \node[neuron] (I-\name) at (0,-\y) {$u_{\y}$};

    \foreach \name / \y in {1,...,3}
    \path[yshift=-0.5cm]
    node[neuron] (H-\name) at (\layersep,-\y cm) {$v_{\y}$};

    \foreach \dest in {1,...,3}{
    \pgfmathtruncatemacro{\destplusone}{\dest + 1};
    \path (I-\dest) edge node[midway,above]{\tiny $w^{*}_1$} (H-\dest);
    \path (I-\destplusone) edge node[midway,above]{\tiny $w^{*}_2$} (H-\dest);
    }

    \node[neuron, fill=red!50] (I-w) at (0,-5) {$u_5$};
    \node[neuron, fill=red!50] (H-w) at (\layersep,-4.5) {$v_4$};
    \path (I-4) edge node[midway,above]{\tiny $w^{*}_1$} (H-w);
    \path (I-w) edge node[midway,above]{\tiny $w^{*}_2$} (H-w);


    \node (matrix) at (7,-2.5) {
      $\mathbf W = 
      \begin{pmatrix}
        w^{*}_{1} & w^{*}_{2} & 0 & 0 & \textcolor{red!50}{0} \\
        0 & w^{*}_{1} & w^{*}_{2} & 0 & \textcolor{red!50}{0} \\
        0 & 0 & w^{*}_{1} & w^{*}_{2} & \textcolor{red!50}{0} \\
        \textcolor{red!50}{0} & \textcolor{red!50}{0} & \textcolor{red!50}{0} & \textcolor{red!50}{w^{*}_{1}} & \textcolor{red!50}{w^{*}_{2}}
      \end{pmatrix}$

    };
  \end{tikzpicture}
  \caption{1D example of the artificial expansion and its associated matrix modification.
    Weights $w^{*}_1, w^{*}_2$ are already trained. The expanded matrix can be obtained by appending
    an additional row and column.
  }
  \label{fig:conv_fake}
  \end{subfigure}
  \caption{Examples of artificially expanding the input array to obtain a larger output.}
  \label{fig:fake}
\end{figure}
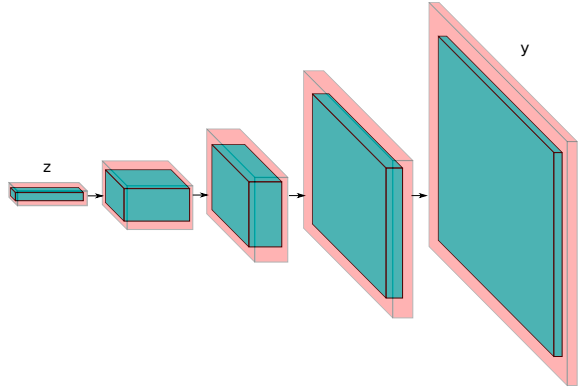
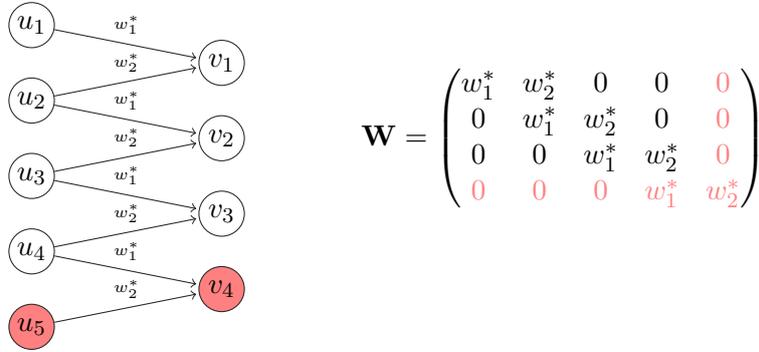

\begin{figure}
  \centering
  \begin{subfigure}{.32\textwidth}
    \includegraphics[width=\textwidth]{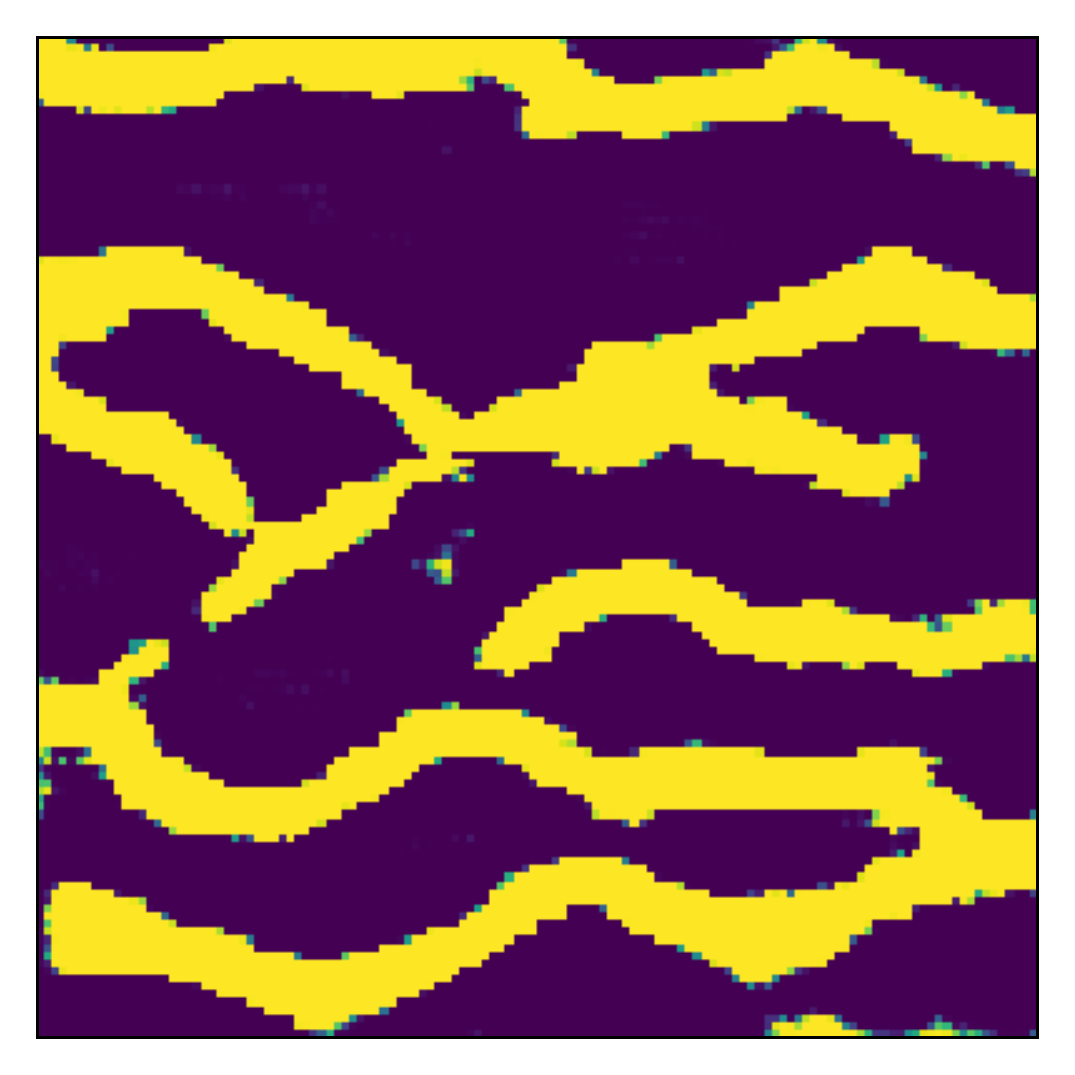}
    \caption{$128\times128$, $n_x=5$}
  \end{subfigure}
  \begin{subfigure}{.32\textwidth}
    \includegraphics[width=\textwidth]{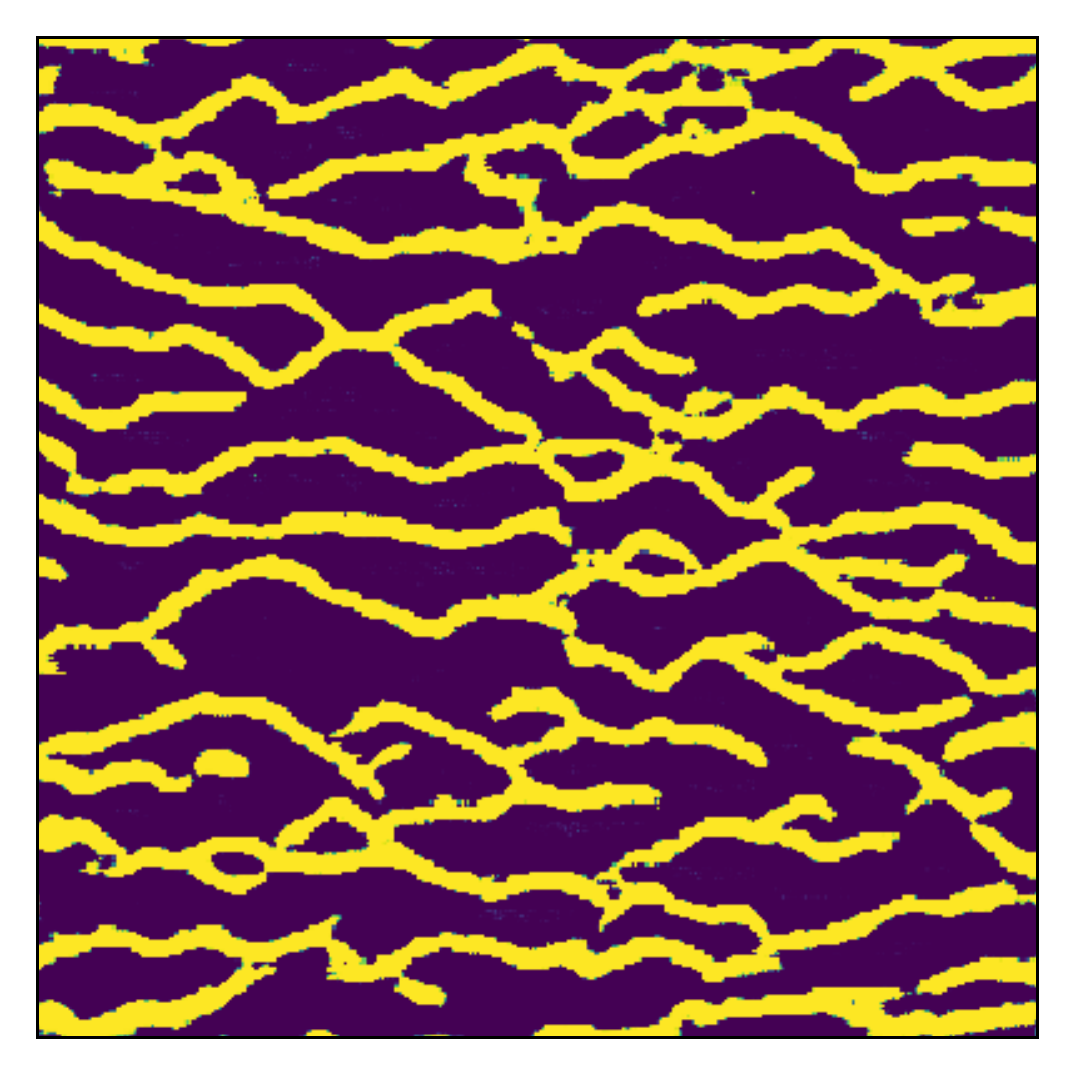}
    \caption{$368\times368$, $n_x=20$}
  \end{subfigure}
  \begin{subfigure}{.32\textwidth}
    \includegraphics[width=\textwidth]{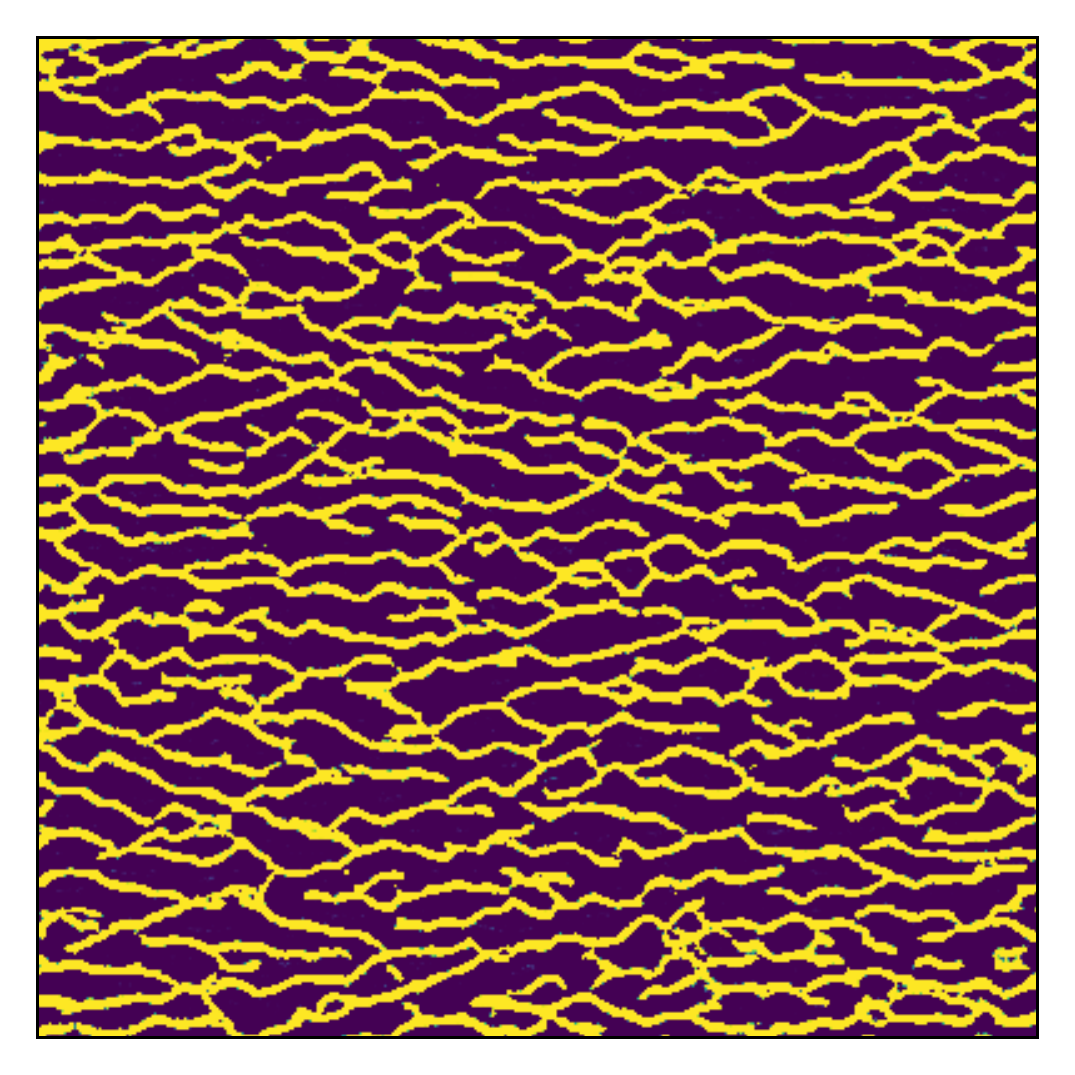}
    \caption{$848\times848$, $n_x=50$}
  \end{subfigure}
  \caption{Artificially upscaled realizations by feeding an expanded input
    array. Images (evidently) not at scale.}
  \label{fig:fake}
\end{figure}

In the domain of geology, a natural question is whether GANs can be directly applied
as multipoint geostatistical simulators. This has been studied in
a number of recent
works~\citep{mosser2017reconstruction,mosser2017stochastic,laloy2018training}.
The idea here is to use a single large training image and simply train a GAN model
on patches of this image, instead of generating a dataset of realizations
using an external multipoint geostatistical simulator.
The result is a generator capable of generating
patches of the image instead of the full-size training image. 
In order to recover the original size of the training image or to generate larger images,
a simple trick is to feed an artificially expanded input array to the generator.
This is illustrated
in~\Cref{fig:cnn_fake}: If a generator has been trained with $\zz$ of shape
$(n_z,1,1)$, we can feed the generator with expanded arrays of shape
$(n_z,n_y,n_x)$ (sampled from the higher dimensional analogue of the same
distribution) to obtain larger outputs. This is
possible since we can still apply a convolving filter regardless of the input
size. This is better illustrated with the 1D example shown
in~\Cref{fig:conv_fake}. In~\Cref{fig:fake}, we show generated examples using this
trick on our trained WGAN model, where we generate realizations of more than
10 times larger in each dimension (recall that our generator is trained on
realizations of size $64\times 64$).
Whether this trick generalizes to arbitrarily large sizes (in the sense that
the flow statistics are preserved) deserves further study.

An important feature of multipoint geostatistical simulators is the ability to generate conditional realizations. Performing conditioning using unconditionally trained generators has been the focus of recent works. In~\citep{dupont2018generating,mosser2018conditioning} conditioning was performed by optimization in the latent space using an image inpainting technique~\citep{yeh2016semantic}.
In~\citep{laloy2018training} conditioning is imposed as an additional constraint in the inversion process.
In our concurrent work~\citep{chan2018parametric}, we propose a method based on stacking a second neural network to the generator that performs the conditioning.

%% file: conclusion.tex
\section{Conclusions}\label{sec:ch3conclusion}

We investigated generative adversarial networks (GAN) as a sample-based parametrization method of stochastic inputs in partial differential equations. We focused on parametrization of geological models which is critical in the performance of subsurface simulations. We parametrized conditional and unconditional permeability, and used the parametrization to perform uncertainty quantification and parameter estimation (history matching). Overall, the method shows very good results in reproducing the spatial statistics and flow responses, as well as preserving visual realism while achieving a dimensionality reduction of two orders of magnitude, from $64\times64$ to $30$. 
In uncertainty quantification, we found that the method reproduces the high order statistics of the flow responses as seen from the estimated distributions of the saturation and the production. The estimated distributions show very good agreement and the modality of the distributions are reproduced.
In parameter estimation, we found successful inversion results in both conditional and unconditional settings, and reasonable inversion results for challenging hand-crafted images. We also compared implementations of the standard formulation of GAN with the Wasserstein formulation, finding the latter to be more suitable for our applications. We discussed issues regarding network size under limited data, highlighting the importance of the choice of discriminator size to prevent overfitting. Finally, we also discussed using GANs for multipoint geostatistical simulations. 
Possible directions to extend this work include improving current GAN methods for limited data, and further assessments in other test cases.